\documentclass[journal]{IEEEtran}

\usepackage{graphicx}
\usepackage[T1]{fontenc}
\usepackage{color}

\usepackage{multirow}
\usepackage{ulem}
\usepackage{amsthm,amsmath,amssymb}
\usepackage{mathrsfs}
\usepackage{url}
\usepackage{subfigure}
\usepackage{booktabs}
\usepackage[driverfallback=dvipdfm,
CJKbookmarks=true, bookmarksnumbered=true, bookmarksopen=true,
colorlinks=true,
pdfborder=001,
citecolor=blue, linkcolor=blue, anchorcolor=red, urlcolor=black
]{hyperref}

\begin{document}
\title{Disentangle Before Anonymize: A Two-stage Framework for Attribute-preserved and Occlusion-robust De-identification}

\author{Mingrui~Zhu,
        Dongxin~Chen,
        Xin~Wei,
        Nannan~Wang,~\IEEEmembership{Senior Member,~IEEE,} and~Xinbo~Gao,~\IEEEmembership{Fellow,~IEEE}
\thanks{}}

\maketitle

\begin{abstract}

In an era where personal photos are easily leaked and collected, face de-identification is a crucial method for protecting identity privacy. However, current face de-identification techniques face challenges in preserving attribute details and often produce anonymized results with reduced authenticity. These shortcomings are particularly evident when handling occlusions, frequently resulting in noticeable editing artifacts. Our primary finding in this work is that simultaneous training of identity disentanglement and anonymization hinders their respective effectiveness. Therefore, we propose ``\textbf{D}isentangle \textbf{B}efore \textbf{A}nonymize'', a novel two-stage \textbf{F}ramework (DBAF) designed for attribute-preserved and occlusion-robust de-identification. This framework includes a \textbf{C}ontrastive \textbf{I}dentity \textbf{D}isentanglement (CID) module and a \textbf{K}ey-authorized \textbf{R}eversible \textbf{I}dentity \textbf{A}nonymization (KRIA) module, achieving faithful attribute preservation and high-quality identity anonymization edits. Additionally, we introduce a \textbf{M}ulti-scale \textbf{A}ttentional \textbf{A}ttribute \textbf{R}etention (MAAR) module to address the issue of reduced anonymization quality under occlusions. Extensive experiments demonstrate that our method outperforms state-of-the-art de-identification approaches, delivering superior quality, enhanced detail fidelity, improved attribute preservation performance, and greater robustness to occlusions.

\end{abstract}


\IEEEpeerreviewmaketitle

\section{Introduction}

\IEEEPARstart{L}{arge} collections of facial images gathered from cameras and social networks are susceptible to malicious access, posing a threat to individual privacy~\cite{meden2021privacy, chen2024fibnet, li2023privacy}. Given the pressing demand for safeguarding identities, the technique of face de-identification has surfaced as a dependable and secure solution. In the early stages, the scope of face de-identification was limited to using pixel-level editing techniques such as adding noise, blurring, masking, \textit{etc.}, to spoil identity information. Despite these methods being able to protect identity privacy to some extent, they directly compromise the usability of the image, resulting in anonymized images that exhibit noticeable editing traces~\cite{vishwamitra2017blur}.

\begin{figure}[!t]
\centering
\includegraphics[width=1.0\columnwidth]{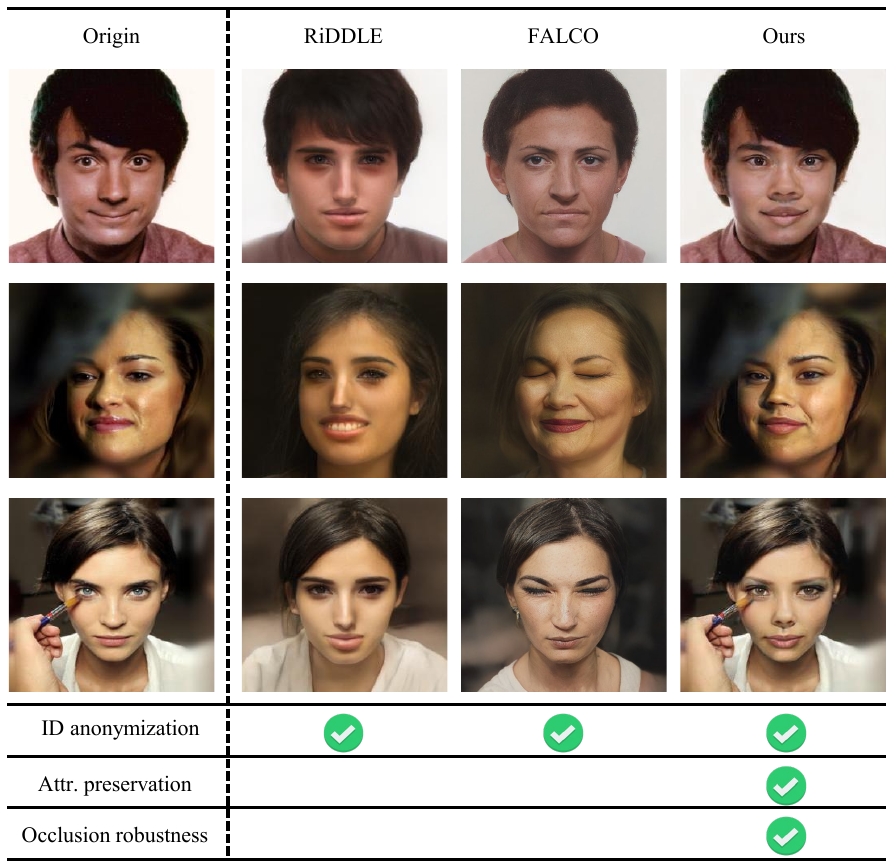}
\caption{De-identification and recovery results generated by RiDDLE~\cite{li2023riddle}, FALCO~\cite{barattin2023attribute}, and our method. Besides being able to synthesize high-fidelity anonymized images, DBAF has obvious advantages in attribute preservation and occlusion robustness.}
\label{Fig1}
\end{figure}

With the advancement of image generation and editing technologies, GAN-based method \cite{hukkelaas2019deepprivacy, maximov2020ciagan, gafni2019live} has made substantial progress in face de-identification. These methods frame the task of face de-identification as a conditional image generation problem, generating anonymized face images with identical attributes but distinct identities compared to the input image used as the condition. Despite demonstrating promising performance, these methods are still susceptible to artifacts, revealing the falsified nature of image edits. Furthermore, many approaches overlook the preservation of attribute information from the original image, including background and brightness characteristics. This oversight results in significant discrepancies between the attributes of the generated results and those of the actual attributes in the original image. Furthermore, most existing methods focus on anonymization, overlooking the reversibility of the anonymization process. When sharing photos online, we aim to conceal identity information, yet we also desire that authorized users can recover the original photo information from the anonymized photos through authentication such as a key.

Recognizing the criticality of reversible de-identification, recent advancements in this field have surfaced~\cite{zhang2023reversible}. FIT~\cite {gu2020password} uses basic networks and complex loss functions to achieve anonymity, but the quality and robustness of its anonymous results are limited. Personal~\cite{cao2021personalized}, on the other hand, promotes personalized manual encryption rules for anonymity, sacrificing network flexibility and security. In contrast, RiDDLE~\cite{li2023riddle} introduces a clear anonymization strategy that significantly enhances generation outcomes. However, it confronts the inherent distortion challenge of StyleGAN~\cite{karras2019style, karras2020analyzing}, where background and facial details deviate noticeably from the original image. RiDDLE addresses this by leveraging a face parsing model to preserve attributes, though its effectiveness in unconventional scenarios like occlusion is somewhat diminished.

After extensive analysis, we believe that the ideal face de-identification method should encompass the following attributes: 1) The ability to generate diverse and lifelike de-identified faces; 2) Efficient restoration of the original identity with utmost reliability; 3) Creation of new identities to ensure privacy protection in insecure environments; 4) Preservation of identity-independent attributes while producing high-quality de-identified faces; and 5) Consistent production of high-quality anonymized results, even in the presence of occlusions.

In this paper, we propose ``\textbf{D}isentangle \textbf{B}efore \textbf{A}nonymize'', a novel two-stage \textbf{F}ramework (DBAF), meeting the above characteristics. The core idea of DBAF is to decouple identity disentanglement and anonymization into two separate stages, in order to enhance the performance of each. Concretely, DBAF comprises a \textbf{C}ontrastive \textbf{I}dentity \textbf{D}isentanglement (CID) module and a \textbf{K}ey-authorized \textbf{R}eversible \textbf{I}dentity \textbf{A}nonymization (KRIA) module. In the first stage, the CID module achieves disentanglement of identity and attributes by separating latent codes into identity-related and attribute-related components, allowing the network to preserve attributes while modifying only the identity. In the second stage, the KRIA module introduces a key into the identity-related code. This key modulates and alters the identity-related code, thereby changing the identity of the final reconstructed image to achieve anonymity. Notably, within this module, we use loss constraints to ensure that the anonymous results controlled by different keys are diverse. Additionally, the original identity image can be restored using the same key, ensuring reversibility. Furthermore, to further preserve attribute-related details and enhance the network's robustness against occlusion interference, we introduce a \textbf{M}ulti-scale \textbf{A}ttentional \textbf{A}ttribute \textbf{R}etention (MAAR) module, which participates in training during both stages of DBAF. Compared to previous methods, our approach enables reversible de-identification while delivering results with higher fidelity and naturalness. In addition, it effectively enhances attribute preservation and occlusion robustness. Fig.~\ref{Fig1} demonstrates the advantages of our approach over state-of-the-art (SotA) methods.

Our contributions can be summarized as follows:
1) We propose a novel ``disentangle before anonymize'' framework that decouples identity disentanglement and anonymization into two separate stages to enhance performance.
2) We design a contrastive identity disentanglement module and a key-authorized reversible identity anonymization module to achieve effective attribute preservation and identity anonymization.
3) We introduce a multi-scale attention-based attribute retention module to further preserve attribute-related details and improve robustness against occlusion.
4) Experimental results show that our approach outperforms prior methods in both de-identification and recovery quality, preserving attribute details more effectively and demonstrating enhanced robustness to occlusion on publicly available datasets.

\section{Related Work}

\subsection{Face De-identification}

Face de-identification involves changing or hiding the identity of faces in images to protect personal privacy from attackers. Traditional face de-identification methods include blurring, masking, introducing noise, and applying pixelation. While these methods can achieve face de-identification, their drawback lies in the significant impact on image fidelity, thus reducing their overall efficacy.

The rise and development of Generative Adversarial Networks (GANs) have introduced new technologies, enabling face de-identification by generating high-quality de-identified images. Recently, the focus on face de-identification has increasingly shifted towards altering an individual's identity while preserving identity-independent attributes. Several approaches~\cite{gafni2019live, wu2018privacy, barattin2023attribute, li2023privacy} aim to optimize networks or latent codes for de-identification by minimizing the cosine similarity of identity features. For instance, Gafni \textit{et al.}~\cite{gafni2019live} introduced a multi-level perceptual loss for de-identification, while Wu \textit{et al.}~\cite{wu2018privacy} generated anonymized images by maximizing differences in identity features. Additionally, the disentangled latent space, known as $\mathcal{W}$~\cite{yang2021semantic, collins2020editing, shen2020interpreting}, has been shown to offer control and editing capabilities. FALCO~\cite{barattin2023attribute} suggests that latent codes in layers 3-7 within the $\mathcal{W}$ space are associated with face identity and optimized these layers' latent codes to achieve anonymization under specific loss constraints. However, this optimization process increases inference time, affecting practicality, and does not explicitly disentangle identities from other attributes, potentially retaining previous identity information in the generated results. Other methods~\cite{hukkelaas2019deepprivacy, maximov2020ciagan, ren2018learning, sun2018natural, sun2018hybrid, hukkelaas2023deepprivacy2, yuan2024pro} attempt to conceal facial identities and synthesize new faces based on provided attribute information. DeepPrivacy~\cite{hukkelaas2019deepprivacy} conditionally generates anonymized images that align with facial context and sparse pose data, while CIAGAN~\cite{maximov2020ciagan} aims to produce anonymized faces using landmarks, masks, and desired identities. Despite these advances, the generated faces often suffer from issues of unnatural appearance, limited diversity, and reduced practicality.

Recently, emerging reversible face de-identification technologies have garnered significant attention~\cite{cao2021personalized, gu2020password, li2023riddle, ye2024securereid, yang2024g}. FIT~\cite{gu2020password} employs a predetermined binary password in conjunction with a face image to guide a generative adversarial network in achieving anonymity through encryption and recovering the original face using the reverse password. Personal~\cite{cao2021personalized} modifies identity characteristics using customized mathematical formulas to enable both anonymization and recovery. RiDDLE~\cite{li2023riddle} introduces a latent encryptor that facilitates anonymity and recovery, with the password serving as a crucial piece of information for decrypting other modes within the latent cipher.
Proença~\cite{proencca2021uu} proposed a reversible face de-identification method specifically designed for video surveillance data. While this method is effective for low-quality scene portraits, it is less suitable for scenarios involving fine-grained facial images. Yuan \textit{et al.}~\cite{yuan2023invertible} introduced an innovative approach to face privacy protection through an invertible image obfuscation framework based on flow, offering a new paradigm in this domain. Wen \textit{et al.}~\cite{wen2022identitymask} developed a modular architecture for reversible face video de-identification, known as IdentityMask, which utilizes deep motion flow to eliminate the need for per-frame evaluation. Nonetheless, this approach prioritizes continuity between images and does not adequately address challenges posed by occluded faces.

\subsection{Face Identity Disentanglement}

Face identity disentanglement focuses on separating identity and attribute representations in facial images to preserve or manipulate identity while maintaining other facial characteristics. Identity-related disentanglement tasks often utilize pre-trained face recognition networks to extract identity features, which guide the training of the disentanglement network. For face swapping, Bao \textit{et al.}~\cite{bao2018towards} developed a simplified disentanglement network using an asymmetric loss to separate identities from attributes. Li \textit{et al.}~\cite{li2021identity} introduced an advanced generator that adaptively regulates the disentangled representations of identity and attributes. Nitzan \textit{et al.}~\cite{nitzan2020face} proposed a StyleGAN-based framework that disentangles facial identities and attributes within the $\mathcal{Z}$ feature space. Similarly, Luo \textit{et al.}~\cite{luo2022styleface} suggested a novel disentanglement framework using StyleGAN, incorporating an adaptive attribute extractor to preserve identity-independent features. In the realm of facial identity preservation, Shen \textit{et al.}~\cite{shen2018faceid} advocated for employing a facial identity classifier as a third-party component to facilitate the separation of identity and attribute features. Shoshan \textit{et al.}~\cite{shoshan2021gan} utilized contrastive learning to achieve GANs with a well-disentangled latent space. Li \textit{et al.}~\cite{li2021identity} designed a region discovery module capable of identifying identity-independent facial attributes, thus enabling the adaptive generation of privacy-preserving faces.

\subsection{GAN Inversion}

GAN inversion maps an image into a disentangled latent space, enabling precise identity manipulation. Several techniques for GAN inversion are currently available, with StyleGAN~\cite{karras2019style, karras2020analyzing} standing out for its exceptional resolution and realistic results, making it the top choice for most inversion tasks. Optimization-driven techniques~\cite{abdal2019image2stylegan, abdal2020image2stylegan++, zhu2020domain} treat the latent code as an optimization target, systematically refining it through iterative processes and loss constraints to ensure the generated output meets the desired specifications. While these methods achieve high-quality results in reconstruction and editing, they are resource-intensive and time-consuming during inference. On the other hand, encoder-based techniques~\cite{alaluf2021restyle, richardson2021encoding, tov2021designing} employ specially crafted encoders to map the input image directly into the $\mathcal{W}$ latent space, streamlining the inference process but with some reduction in fidelity compared to optimization-driven approaches. Considering the trade-off between fidelity and efficiency, we opted for an encoder-based inversion approach~\cite{tov2021designing}, which strikes a practical balance between speed and quality.

\section{Method}

\subsection{Preliminary}

\begin{figure}[!t]
\centering
\includegraphics[width=1.0\columnwidth]{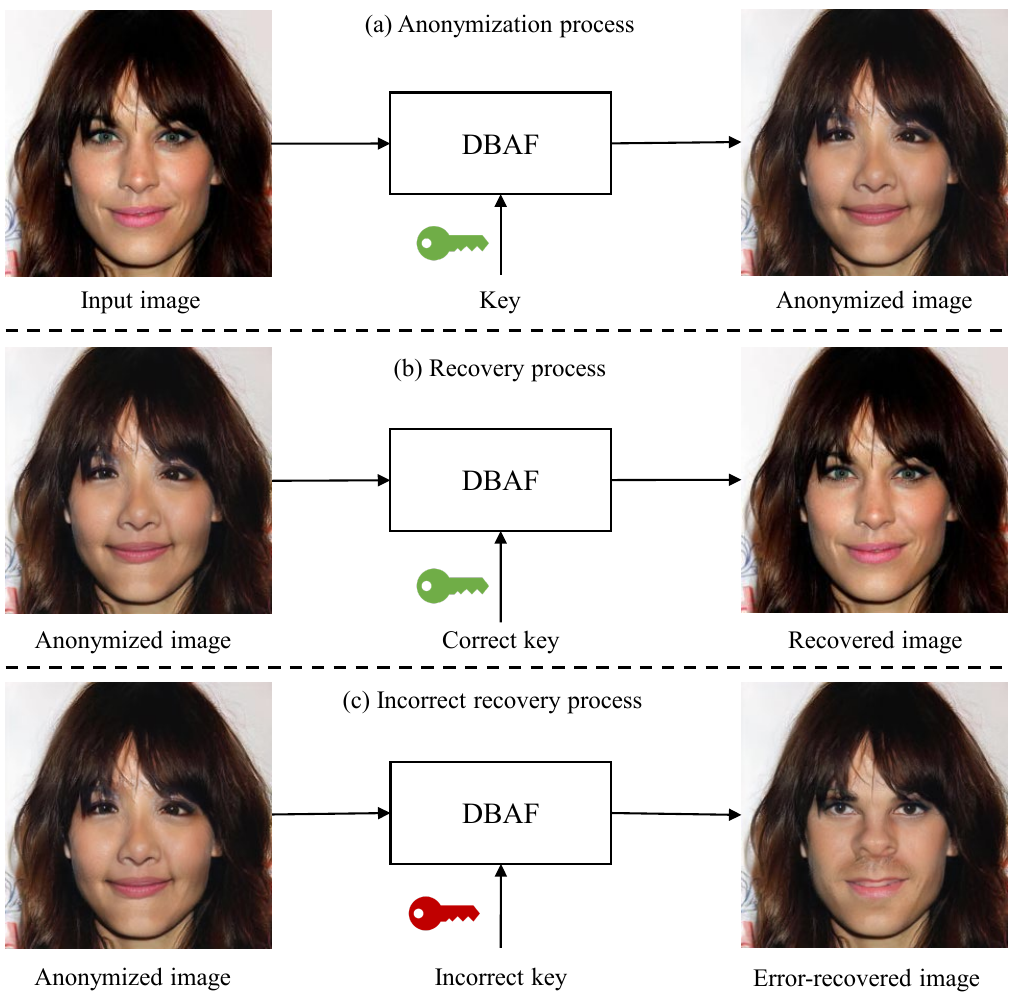}
\caption{The anonymization and recovery process of DBAF. In the anonymization phase, given the input face image $X_{ori}$ and a randomly generated key $P$, DBAF transforms the input face into an anonymized version. During the recovery phase, if the entered key matches the one generated in the anonymization process, DBAF generates an image with the original identity. Conversely, if the entered key does not match, DBAF produces an image with a different identity.}
\label{Fig2}
\end{figure}

The objective is to anonymize a given face image by altering its identity information while accurately preserving identity-irrelevant attributes such as hairstyle, facial expressions, and background. At the same time, the process ensures that the original face can be reliably reconstructed from the anonymized version. A key is utilized to control both the face anonymization and recovery processes. The original face can only be accurately reconstructed by providing the exact key that was used during the anonymization. Formally, let $X_{ori}$ and $X_{ano}$ denote the input face image and its anonymized counterpart, respectively. The recovered original image from $X_{ano}$ is denoted as $X_{rec}$. In the anonymization process, given the input face image $X_{ori}$ and a randomly generated key $P$, our objective is to learn a mapping $\mathcal{G}(\cdot)$ that transforms the input face into an anonymized version: $X_{ano} = \mathcal{G}(X_{ori}, P)$. During the recovery process, the goal is to reconstruct the original input face $X_{rec}$ from the anonymized face $X_{ano}$ using the correct key $P$ with the same mapping function: $X_{rec} = \mathcal{G}(X_{ano}, P)$. If an incorrect key is provided, the original face cannot be recovered, and the resulting reconstructed face will possess a completely different identity. Fig.~\ref{Fig2} illustrates the anonymization and recovery process of DBAF.

\subsection{Disentangle Before Anoymize}

To achieve the above goals, cutting-edge methods modify identity-related latent codes in the $\mathcal{W}$ space of StyleGAN~\cite{karras2019style} and ensure the decoupling of identity and attributes through loss constraints. In this paradigm, however, the model is tasked with achieving disentanglement and anonymization objectives simultaneously, which hinders their respective effectiveness. Moreover, the $\mathcal{W}$ space of StyleGAN cannot ensure a complete disentanglement between identity and identity-unrelated attributes in anonymization tasks. As a result, the identity-unrelated attribute information in the anonymized images is unfortunately altered, and this issue becomes more severe in scenarios involving occlusions. Therefore, we propose a novel ``Disentangle Before Anonymize'' Framework (DBAF) for attribute-preserved and occlusion-robust face de-identification. The core idea of this framework is to prioritize disentanglement before anonymization, which provides two significant advantages: first, achieving a more compact disentangled space ensures a clear separation between identity and attributes, resulting in a ``cleaner'' identity latent code for the anonymization process; second, splitting the overall objective into two distinct stages allows each stage to focus exclusively on its specific task, thereby improving the effectiveness and efficiency of both stages.

\begin{figure}[!t]
\centering
\includegraphics[width=1.0\columnwidth]{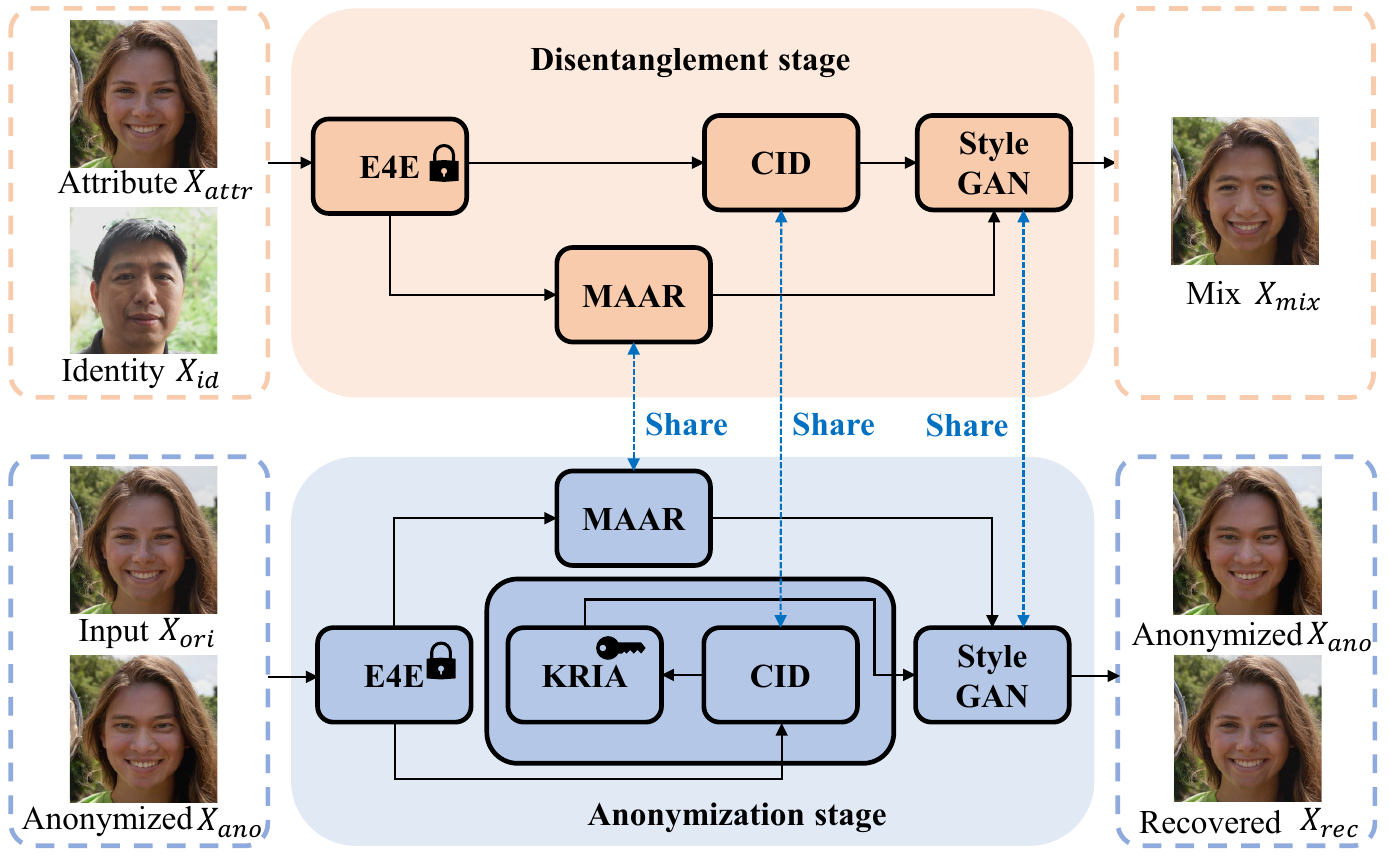}
\caption{Overview of the stages and modular structure of DBFA. On the basis of a pre-trained E4E encoder~\cite{tov2021designing} and a StyleGAN2 decoder~\cite{karras2020analyzing}, DBAF introduces three modules: CID, KRIA, and MAAR. CID and MAAR are involved in both stages of training, whereas KRIA is trained exclusively in the second stage. The two stages utilize different inputs and loss constraints to guide the model's learning effectively.}
\label{Fig3}
\end{figure}

Fig.~\ref{Fig3} provides an overview of the stages and modular structure of our proposed method. Concretely, building on a pre-trained E4E encoder~\cite{tov2021designing} and a StyleGAN2 decoder~\cite{karras2020analyzing}, DBAF introduces three modules: the Contrastive Identity Disentanglement (CID) module, the Key-Authorized Reversible Identity Anonymization (KRIA) module, and the Multi-Scale Attentional Attribute Retention (MAAR) module. These modules are involved in the training of two stages: the disentanglement stage and the anonymization stage. 

In the disentanglement stage, the input consists of a pair of images: one provides attribute information, denoted as $X_{attr}$, and the other provides identity information, denoted as $X_{id}$. The E4E encoder transforms each image into a latent code along with corresponding spatial feature maps. Then, CID disentangles and recombines the identity latent code from the identity image $X_{id}$ with the attribute latent code derived from the attribute image $X_{attr}$. Simultaneously, MAAR enhances the attribute details within the spatial features of $X_{attr}$. The recombined latent code and enhanced spatial features are fed into the StyleGAN decoder to generate a mixed image, $X_{mix}$, which seamlessly combines the identity from the identity image $X_{id}$ with the attributes from the attribute image $X_{attr}$.

In the anonymization stage, the input is an image to be anonymized, $X_{ori}$. Similar to the first stage, the input image is encoded into a latent code and spatial feature maps using the E4E encoder. Subsequently, CID disentangles the latent code of the input image into an identity latent code and an attribute latent code. A newly introduced module, KRIA, enables the editing of the identity latent code through key-based control. The edited identity latent code is then recombined with the unaltered attribute latent code. Meanwhile, MAAR is employed to enhance the attribute details within the spatial features of the input image. The recombined latent code and enhanced spatial features are input into the StyleGAN decoder to generate an anonymized image, $X_{ano}$, which preserves the attributes of the input image while exhibiting a distinctly different identity. Additionally, to achieve reversible anonymization, the anonymized image, $X_{ano}$, is also used as input at this stage. By utilizing the same key that was applied during the anonymization process to control identity transformation, the recovered image, $X_{rec}$, is ensured to be identical to the original image, $X_{ori}$.

Note that in both stages, the training of the model is guided by different loss constraints, ensuring that the three proposed modules achieve their respective functionalities. During testing, we directly use the complete model from the second stage to perform anonymization or recovery of the input image.

\begin{figure*}[!t]
\centering
\includegraphics[width=2.0\columnwidth]{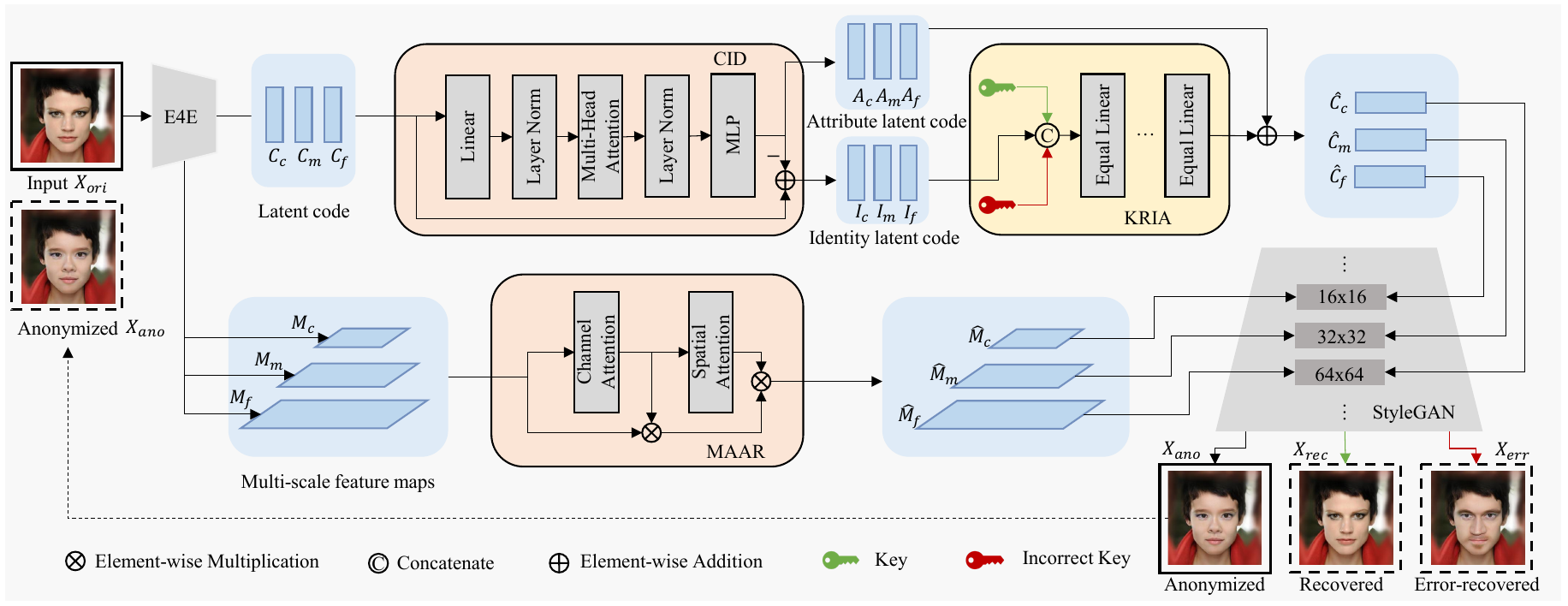}
\caption{The architecture of DBAF. DBAF comprises three key components. The CID module separates the latent code of the input image into an attribute latent code and an identity latent code. The KRIA module edits the identity latent code using key-based controls to achieve identity modifications. The MAAR module preserves and enhances the attribute details, ensuring high fidelity and accurate attribute representation.}
\label{Fig4}
\end{figure*}

\subsection{Disentanglement Stage}
Given an attribute image $X_{attr}$ and an identity image $X_{id}$, we first adopt the E4E encoder~\cite{tov2021designing} to extract their respective latent codes and spatial feature maps at different levels. Specifically, we extract latent codes at three different levels: $C_c\in R^{4\times d}$, $C_m\in R^{4\times d}$, $C_f\in R^{6\times d}$, corresponding to the coarse, medium, and fine levels, respectively. The spatial feature maps at different resolutions are denoted as $M_{k_1}\in R^{k_1\times k_1\times d_1}$, $M_{k_2}\in R^{k_2\times k_2\times d_2}$, and $M_{k_3}\in R^{k_3\times k_3\times d_3}$.

In the CID module, the latent codes $C_c\in R^{4\times d}$, $C_m\in R^{4\times d}$, and $C_f\in R^{6\times d}$ at three levels are processed independently by a shared Multi-head Attention (MHA) module to extract information across different levels. For example, consider the coarse-level latent code $C_c$. In the self-attention layer, the queries, keys, and values are obtained by linearly projecting the view features. Specifically, the query $Q$, key $K$, and value $V$ are denoted as: 
\begin{equation}
Q \triangleq C_cW^Q, \quad K \triangleq C_cW^K, \quad V \triangleq C_cW^V,
\end{equation}
where $W^Q \in \mathbb{R}^{d \times d}$, $W^K \in \mathbb{R}^{d \times d}$, and $W^V \in \mathbb{R}^{d \times d}$ are learnable linear weights. We utilize the Scaled Dot-Product Attention defined as:
\begin{equation}
\text{Attention}(Q,K,V) = \text{softmax}(\frac{QK^T}{\sqrt{d_k}})V.
\end{equation}
Then the MHA is calculated as: 
\begin{align}\label{eq:MHA} 
\begin{split}
\text{MHA}(Q,K,V) &= \text{Concat}(head_1, ..., head_h)W^O, \\
\text{where \:}head_h &= \text{Attention}(Q, K, V).
\end{split}
\end{align}
Here, $W^o\in \mathbb{R}^{hd \times d}$ reduces the dimension of the concatenated attention heads. The relationships among the latent codes at different levels are explored, allowing us to derive the attribute latent codes $A_c$, $A_m$, and $A_f$ at coarse, medium, and fine levels. Furthermore, the identity latent codes $I_c$, $I_m$, and $I_f$ are obtained by adding the latent codes $C_c$, $C_m$, and $C_f$ to the corresponding attribute latent codes $A_c$, $A_m$, and $A_f$, respectively: 
\begin{equation}
I_c=C_c-A_c, I_m=C_m-A_m, I_f=C_f-A_f.
\end{equation}
The underlying idea behind this operation is to treat the attribute latent code as a residual and disentangle the identity latent code from the attribute latent code. We then recombine the attribute latent codes $A_c^{attr}$, $A_m^{attr}$, and $A_f^{attr}$ from the attribute image $X_{attr}$ with the identity latent codes $I_c^{id}$, $I_m^{id}$, and $I_f^{id}$ from the identity image $X_{id}$ to obtain the reconstructed latend codes:
\begin{equation}
\hat{C}_c=I_c^{id}+A_c^{attr}, \hat{C}_m=I_m^{id}+A_m^{attr}, \hat{C}_f=I_f^{id}+A_f^{attr}.
\end{equation}

The MAAR module is designed to select the spatial features most relevant to the given attribute. To achieve this, we apply both channel attention and spatial attention to the spatial feature maps. First, we compute the channel attention weights:
\begin{equation}
{M}^{'}_{k_i} = \Phi_{CA}(\Psi_{CME}(M_{k_i}W_{k_i})+\Psi_{CAE}(M_{k_i}W_{k_i})),
\end{equation}
where ${M}^{'}_{k_i}\in R^{k_i\times k_i \times d}$ represents the channel attention weights, which have the same dimensions as the spatial feature map $M_{k_i}$. The learnable parameters $W_{k_i}\in R^{d_i\times d}$ expand the number of channels in the feature map to $d$. $\Psi_{CME}$ and $\Psi_{CAE}$ are channel-wise max-pooling and channel-wise average-pooling operations, respectly, followed by a dimensional expansion that repeats the pooled matrix along the channel dimension, resulting in an output of size $k_i\times k_i \times d$. $\Phi_{CA}$ is a fusion function implemented as a one-layer MLP with $d$ hidden unit and a sigmoid activation function. Next, we generate the spatial attention weights by:
\begin{equation}
\overline{M}_{k_i} = \Phi_{SA}([\Psi_{SME}({M}^{'}_{k_i}),\Psi_{SAE}({M}^{'}_{k_i})]),
\end{equation}
where $\Psi_{SME}$ and $\Psi_{SAE}$ represent spatial-wise max-pooling and average-pooling operations, respectively. These operations are followed by a dimensional expansion step, where the pooled matrices are repeated along the spatial dimension, resulting in an output of size $k_i\times k_i \times d$. The notation $[\cdot]$ denotes the concatenation of two vectors along the channel dimension. Additionally, $\Phi_{SA}$ is implemented as a single-layer multi-layer perceptron with $d$ hidden units and a sigmoid activation function. Finally, the output of the MAAR module is computed as the element-wise (Hadamard) product of the spatial feature maps with their corresponding channel attention weights and spatial attention weights:
\begin{equation}
\hat{M}_{k_i}= \overline{M}_{k_i} \cdot{M}^{'}_{k_i}\cdot(M_{k_i}W_{k_i}).
\end{equation}  

The reconstructed latent codes from the CID module, along with the enhanced spatial features from the MAAR module, are input into the StyleGAN decoder to generate a mixed image $X_{mix}$:
\begin{equation}
    X_{mix}=\phi_{Style}(\hat{M}_c^{attr},\hat{M}_m^{attr},\hat{M}_f^{attr}, \hat{C}_c^{attr},\hat{C}_m^{attr},\hat{C}_f^{attr}),
\end{equation}
where $\phi_{Style}$ represents the internal computational logic of StyleGAN.

\noindent\textbf{Training Objective.} To achieve disentanglement, we propose a specially designed contrastive loss function:
\begin{equation}
L_{ctr}= \lambda\cdot\phi_{max}(D_{c} -\tau ^{+},0) +(1-\lambda)\cdot\phi_{max}(\tau ^{-}-D_{c},0), 
\end{equation}
where $\phi_{max}$ denotes a function that returns the maximum value. Specifically, when the identity image $X_{id}$ and the and the attribute image $X_{attr}$ are identical, the threshold is set to $\tau ^{+}$ , and $\lambda$ is assigned a value of 1. Conversely, when $X_{id}$ and $X_{attr}$ are different, the threshold is set to $\tau ^{-}$, and $\lambda$ is assigned a value of 0. The term $D_{c}$ represents the cosine distance between the mixed image $X_{mix}$ and the attribute image $X_{attr}$:
\begin{equation}
D_{c}=1-\frac{F_{arc}(X_{mix} )\cdot F_{arc}(X_{attr} ) }{\left |   F_{arc}(X_{mix} )\right |\cdot \left |F_{arc}(X_{attr} )\right | }, 
\end{equation}
where $F_{arc}$ refers to a pre-trained ArcFace model~\cite{deng2019arcface}, which is utilized for extracting identity embeddings. We calculate the $L_2$ distance between the encoded embedding of $X_{mix}$ and $X_{attr}$ extracted by the LPIPS model~\cite{zhang2018unreasonable} to quantify the perceptual loss between $X_{mix}$ and $X_{attr}$:
\begin{equation}
L_{lpips}=||F_{l}(X_{attr})-F_{l}(X_{mix} )||_{2},         
\end{equation}
where $F_l$ denotes the pre-trained LPIPS model. We compute the  $L_1$ distance between $X_{mix}$ and $X_{attr}$ to quantify the pixel-level reconstruction loss:
\begin{equation}
L_{rec}=||X_{attr}-X_{mix}||_{1}.          
\end{equation}
To ensure the generated facial structure appears natural, we incorporate a face parsing loss:
\begin{equation}
L_{parse}=||F_{p}(X_{attr} )-F_{p}(X_{mix} )||_{2},         
\end{equation}
where $F_{p}$ denotes the pre-trained face parsing model~\cite{lee2020maskgan}. To align the output latent code 
w more closely with the $\mathcal{W}^+$ latent space, we introduce a regularization loss:
\begin{equation}
L_{reg}=\sum_{i=1}^{N-1} ||w_i-\bar{w} ||_{2},       
\end{equation}
where $w_i$ denotes the style code of the output latent codes, $\bar{w}$ represents the averaged latent code, and $N$ specifies the number of layers in the latent code. We incorporate an adversarial loss, complemented by $R_1$ regularization, to improve the realism of the generated images:
\begin{equation}
\begin{aligned}
L_{adv}^{D}&=-\mathbb{E}[\log{D(X_{attr})}]-\mathbb{E}[\log{(1-D(X_{mix}))}] \\
& {+\frac{\gamma }{2}\mathbb{E}[||\bigtriangledown D(X_{attr})||_{2}^{2}] }, \\ 
\end{aligned}
\end{equation}
\begin{equation}
L_{adv}^{G}=-\mathbb{E}[\log{D(X_{mix})}].    
\end{equation}
The discriminator $D$ used in this loss follows a framework similar to the one described in StyleGAN2~~\cite{karras2020analyzing}. The full loss for the disentanglement stage is:
\begin{equation}
\begin{aligned}
L_{s1}&=\lambda _{c}L_{ctr}+\lambda _{lpips}L_{lpips}\\
&+\lambda _{rec}L_{rec}+\lambda _{parse}L_{parse}+\lambda _{reg}L_{reg}+L_{adv}.\\\label{stage1_loss}
\end{aligned}
\end{equation}
The loss hyperparameters are empirically configured as follows: $\lambda _{c}=1.0$, $\lambda _{lpips}=1.0$, $\lambda _{rec}=3.5$, $\lambda _{parse}=0.1$, and $\lambda _{reg}=0.1$.

\subsection{Anoymization Stage}

As show in Fig.~\ref{Fig4}, given the input image $X_{ori}$ and the anonymized image $X_{ano}$, their corresponding attribute latent codes $\{A_z^{ori}\}_{\{z=c,m,f\}}$ and $\{A_z^{ano}\}_{\{z=c,m,f\}}$, as well as identity latent codes $\{I_z^{ori}\}_{\{z=c,m,f\}}$ and $\{I_z^{ano}\}_{\{z=c,m,f\}}$, are produced by the CID module. In the anonymization stage, we further introduce the KRIA module to manipulate the latent codes through key-based control, enabling identity transformation. To achieve reversible and diverse anonymization, KRIA employs four distinct keys to generate five types of reconstructed latent codes:
\begin{equation}
\hat{C}_z^{ano_1} = A_z^{ori}+\phi_{KRIA}([P_z^{cor_1},I_z^{ori}]),
\end{equation}
\begin{equation}
\hat{C}_z^{ano_2} = A_z^{ori}+\phi_{KRIA}([P_z^{cor_2},I_z^{ori}]),
\end{equation}
\begin{equation}
\hat{C}_z^{rec} = A_z^{ano}+\phi_{KRIA}([P_z^{cor_1},I_z^{ano}]),
\end{equation}
\begin{equation}
\hat{C}_z^{err_1} = A_z^{ano}+\phi_{KRIA}([P_z^{err_1},I_z^{ano}]),
\end{equation}
\begin{equation}
\hat{C}_z^{err_2} = A_z^{ano}+\phi_{KRIA}([P_z^{err_2},I_z^{ano}]),
\end{equation}
where $[\cdot, \cdot]$ denotes the concatenation of two vector, and $\phi_{KRIA}$ represents a seven-layer equal-linear block as described in~\cite{abdal2019image2stylegan}. The keys $\{P_z^{cor_1}\}_{\{z=c,m,f\}}$ and $\{P_z^{cor_2}\}_{\{z=c,m,f\}}$ are correct keys, designed to control the generation of two completely distinct anonymized results. Conversely, $\{P_z^{err_1}\}_{\{z=c,m,f\}}$ and $\{P_z^{err_2}\}_{\{z=c,m,f\}}$ are incorrect keys, intended to restore the anonymized image to two error-recovered images that differ from the original image. All keys have the same dimensions $I_z^{ori}$. Simultaneously, the spatial feature maps $\{M_z^{ori}\}_{\{z=c,m,f\}}$ and $\{M_z^{ano}\}_{\{z=c,m,f\}}$, generated by the E4E encoder, processed by the MAAR module to enhance attribute details, resulting in $\{\hat{M}_z^{ori}\}_{\{z=c,m,f\}}$ and $\{\hat{M}_z^{ano}\}_{\{z=c,m,f\}}$.

The reconstructed latent codes, along with the enhanced spatial features, are input into the StyleGAN decoder to generate five types of images:
\begin{equation}
X_{ano_1} = \phi_{style}(\hat{M}_z^{ori}, \hat{C}_z^{ano_1}),
\end{equation}
\begin{equation}
X_{ano_2} = \phi_{style}(\hat{M}_z^{ori}, \hat{C}_z^{ano_2}),
\end{equation}
\begin{equation}
X_{rec} = \phi_{style}(\hat{M}_z^{ano}, \hat{C}_z^{rec}),
\end{equation}
\begin{equation}
X_{err_1} = \phi_{style}(\hat{M}_z^{ano}, \hat{C}_z^{err_1}),
\end{equation}
\begin{equation}
X_{err_2} = \phi_{style}(\hat{M}_z^{ano}, \hat{C}_z^{err_2}),
\end{equation}
where $\phi_{Style}$ represents the internal computational logic of StyleGAN. The outputs $X_{ano_1}$, $X_{ano_2}$, $X_{rec}$, $X_{err_1}$, and $X_{err_2}$ correspond to the anonymized image 1, anonymized image 2, correct-recovered image, error-recovered image 1, and error-recovered image 2, respectively.

\noindent\textbf{Training Objective.} For convenience, we denote the set composed of $X_{ano_1}$, $X_{ano_2}$, $X_{err_1}$, and $X_{err_2}$ as $S_1$, \textit{i.e.}, $S_1 = \{X_{ano_1}, X_{ano_2}, X_{err_1}, X_{err_2}$\}. To ensure that the identities of anonymized images and error-recovered images exhibit significant differences from the identity of the original image, we define the identity difference loss as:
\begin{equation}
L_{dif}= \sum_{\substack{x\in S_{1}}} \phi_{max} (0,\cos(F_{arc}(X_{ori}),F_{arc}(x))),
\end{equation}
where $F_{arc}$ refers to a pre-trained ArcFace model~\cite{deng2019arcface}, which is utilized for extracting identity embeddings. $\phi_{max}$ is a function that returns the maximum value. To ensure that the identity of correct-recovered image aligns with the identity of the original image, we define the identity recovery loss as:
\begin{equation}
L_{rev}= 1-\cos(F_{arc}(X_{ori}),F_{arc}(X_{rec})).  
\end{equation}
To facilitate the diversity of identity changes caused by password differences, we define the identity diversity loss as:
\begin{equation}
L_{div}=\sum_{\substack{x,y\in S_{1}, x\ne y}} \phi_{max} (0,\cos(F_{arc}(x),F_{arc}(y))).
\end{equation}
Therefore, the identity related loss of the anoymization stage is:
\begin{equation}
L_{id} = L_{dif} + L_{rec} + L_{div}.      
\end{equation}
For convenience, we denote the set composed of $X_{ano_1}$, $X_{ano_2}$, $X_{rec}$, $X_{err_1}$, and $X_{err_2}$ as $S_2$, \textit{i.e.}, $S_2 = \{X_{ano_1}, X_{ano_2}, X_{rec}, X_{err_1}, X_{err_2}$\}. To ensure high image quality and the preservation of attribute features, we incorporate the LPIPS loss~\cite{zhang2018unreasonable}, which is defined as:
\begin{equation}
L_{lpips}=\sum_{\substack{x\in S_{2}}} ||F_{l}(X_{ori} )-F_{l}(x)||_{2},          
\end{equation}
where $F_l$ denotes the pre-trained LPIPS model. We also compute
the $L_1$ distance between the generated images and the original image to quantify the pixel-level reconstruction loss: 
\begin{equation}
L_{rec}=\sum_{\substack{x\in S_{2}}} ||X_{ori}-x||_{1}.          
\end{equation}
To ensure the generated facial structure appears natural, we
incorporate a face parsing loss:
\begin{equation}
L_{parse}=\sum_{\substack{x\in S_{2}}} ||F_{p}(X_{ori} )-F_{p}(x )||_{2},          
\end{equation}
where $F_{p}$ denotes the pre-trained face parsing mode~\cite{lee2020maskgan}. To align the output latent code 
w more closely with the $\mathcal{W}^+$ latent space, we introduce a regularization loss:
\begin{equation}
L_{reg}=\sum_{i=1}^{N-1} ||w_i-\bar{w} ||_{2},       
\end{equation}
where $w_i$ denotes the style code of the output latent codes, $\bar{w}$ represents the averaged latent code, and $N$ specifies the number of layers in the latent code. We incorporate an adversarial loss, complemented by $R_1$ regularization, to improve the realism of the generated images:
\begin{equation}
\begin{aligned}
L_{adv}^{D}&=\sum_{\substack{x\in S_{2}}}(-\mathbb{E}[\log{D(X_{ori})}]-\mathbb{E}[\log{(1-D(x))}] \\
& {+\frac{\gamma }{2}\mathbb{E}[||\bigtriangledown D(X_{ori})||_{2}^{2}] }), \\ 
\end{aligned}
\end{equation}
\begin{equation}
L_{adv}^{G}=\sum_{\substack{x\in S_{2}}}-\mathbb{E}[\log{D(x)}].  
\end{equation}
The full loss of the anoymization stage is:
\begin{equation}
\begin{aligned}
L_{s2}&=\lambda _{id}L_{id}+\lambda _{lpips}L_{lpips}\\
&+\lambda _{rec}L_{rec}+\lambda _{parse}L_{parse}+\lambda _{reg}L_{reg}+L_{adv},\\\label{total_loss}
\end{aligned}
\end{equation}
The loss hyperparameters are empirically configured as follows: $\lambda _{id}=2.0$, $\lambda _{lpips}=1.0$, $\lambda _{rec}=0.05$, $\lambda _{parse}=0.1$, and $\lambda _{reg}=0.1$.

\begin{figure*}[!t]
\centering
\includegraphics[width=2.0\columnwidth]{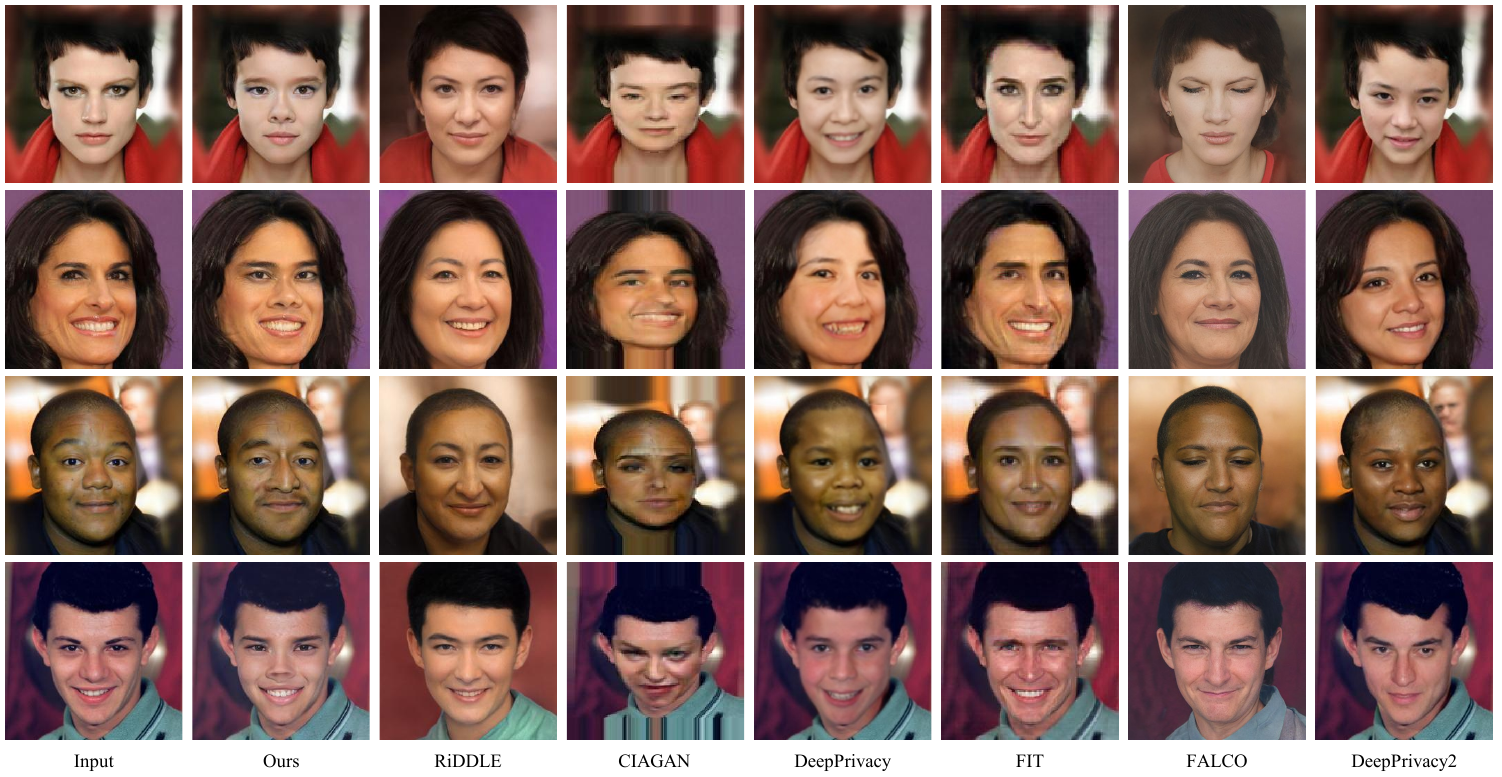}
\caption{Qualitative de-identification results on CelebA-HQ~\cite{karras2017progressive}. Our method produces high-quality images while preserving key attribute features, including background details and fine textures such as wrinkles, more effectively than existing approaches.}
\label{Fig5}
\end{figure*}

\section{Experiments}

\subsection{Experimental Setup}

\noindent\textbf{Implementation details.} Our model is trained on a 48GB NVIDIA Quadro RTX 8000 GPU with a total batch size of 8. Training is performed using the Ranger optimizer with a learning rate of 0.0001. In the anonymization stage, two passwords are used: one correct password and one randomly generated password. For the recovery process, three passwords are used: one correct password and two incorrect ones.

\noindent\textbf{Datasets.} We conduct experiments on the FFHQ~\cite{karras2019style} and CelebA-HQ~\cite{karras2017progressive} datasets. The FFHQ dataset contains 70k high-quality face images, of which we randomly selected 69k for training and reserved the remaining 1k images for testing. Additionally, we randomly selected 1k images from the CelebA-HQ dataset for testing.

\noindent\textbf{Compared Methods.} We compare our method with 6 leading face de-identification methods: RiDDLE~\cite{li2023riddle}, CIAGAN~\cite{maximov2020ciagan}, FIT~\cite{gu2020password}, Deep Privacy~\cite{hukkelaas2019deepprivacy}, Deep Privacy2~\cite{hukkelaas2023deepprivacy2}, and FALCO~\cite{barattin2023attribute}. Among them, FIT~\cite{gu2020password} and RiDDLE~\cite{li2023riddle} are reversible face de-identification methods, capable of achieving both face identity anonymization and recovery, while the other methods can only achieve face identity anonymization.

\noindent\textbf{Evaluation Metrics.} To quantitatively evaluate the anonymized and recovered images, we adopt two evaluation methods: (1) using recognition rate metrics to verify whether the anonymized images and recovered images achieve the goals of anonymization and recovery, and (2) employing image quality evaluation metrics to assess the quality of the synthesized images and their ability to preserve attributes. Specifically, we employ FaceNet~\cite{schroff2015facenet} and ArcFace~\cite{deng2019arcface} to calculate the recognition rate. For quality evaluation, we utilize the Mean Squared Error (MSE) and Peak Signal-to-Noise Ratio (PSNR) to evaluate pixel-level distortion, Learned Perceptual Image Patch Similarity (LPIPS)~\cite{zhang2018unreasonable} metric to measure perceptual similarity, and Frechet Inception Distance (FID)~\cite{heusel2017gans} to quantify the distributional divergence between original and synthesized images.

\subsection{De-identification\label{sec:De-identification}}

We present a qualitative comparative analysis of the proposed method against recent state-of-the-art (SotA) approaches on the CelebA-HQ dataset, as illustrated in Fig.~\ref{Fig5}. While FIT~\cite{gu2020password} excels in generating anonymized faces, its visual quality has room for improvement. CIAGAN~\cite{maximov2020ciagan} often generates unnatural facial features and distorted backgrounds, as seen in all the results generated by CIAGAN. DeepPrivacy~\cite{hukkelaas2019deepprivacy} preserves photo realism to some extent but fails to maintain identity-independent attributes like facial expressions. RiDDLE~\cite{li2023riddle} generates diverse anonymized faces, but certain features appear unnatural, such as the nose in the third row of the illustrated figure. Across RiDDLE-generated results, the faces often lack sufficient realism, with overly smooth textures and an absence of authentic skin details, as shown in the second and third rows. DeepPrivacy2~\cite{hukkelaas2023deepprivacy2} performs well in most scenarios but can introduce facial distortions, as evident in the first and second rows of the figure. FALCO~\cite{barattin2023attribute} leverages the strong generative capabilities of StyleGAN to produce clear de-identified faces but tends to alter facial expressions—e.g., closed eyes in the second and third rows and closed mouths in the first and fourth rows. Furthermore, FALCO often replaces or removes the original background in its outputs. In contrast, our method generates anonymized images with superior fidelity, realism, and robustness to occlusions, while preserving both identity-independent attributes and the original background context.

We evaluate the anonymization performance of various methods by calculating the recognition rate on the CelebA-HQ dataset~\cite{karras2017progressive}. Anonymization is deemed successful when the distance between the de-identified image and the source image in the embedding space of a face recognition model exceeds the threshold defined by the respective recognition network. In this study, we use FaceNet~\cite{schroff2015facenet} and ArcFace~\cite{deng2019arcface} for evaluation, with thresholds set at 1.1 for FaceNet and 0.8 for ArcFace, as recommended in~\cite{schroff2015facenet}. To ensure a fair comparison, we use an equal number of images (200 per method) to evaluate the effectiveness of de-identification. The evaluation results are summarized in TABLE~\ref{table:1}. A lower recognition rate for de-identified images corresponds to better anonymization performance. Our method demonstrates a clear advantage by providing more reliable privacy protection compared to alternative approaches. To evaluate the quality and attribute preservation performance of anonymized images, we calculate the similarity between anonymized images and original images. Higher similarity indicates better fidelity of the anonymized images and superior preservation of identity-independent attributes. As shown in TABLE~\ref{table:2}, our method achieves the best performance across all four metrics (MSE, PSNR, LPIPS, and FID), demonstrating superior synthesis quality and attribute preservation compared to other methods.

\begin{table}[!t]
\caption{
Quantitative evaluation (recognition rate) on CelebA-HQ. According to \cite{schroff2015facenet}, the thresholds for recognition are set as $\tau_{facenet} = 1.1$ for FaceNet and $\tau_{arcface} = 0.8$ for ArcFace. The best results are in \textbf{bold}, and the second best results are \underline{underlined}.}
\begin{tabular}{lcccc}
\toprule
\multicolumn{1}{c}{Type}               & Method        & \begin{tabular}[c]{@{}c@{}}FaceNet\\ CASIA\end{tabular} & \begin{tabular}[c]{@{}c@{}}FaceNet\\ VGGFace2\end{tabular} & ArcFace           \\ \hline
                                       & Ours          & \textbf{0.012}                                          & \textbf{0.02}                                              & \textbf{0}        \\
                                       & RiDDLE        & \underline{0.016}                                       & \underline{0.032}                                          & \textbf{0}        \\
                                       & CIAGAN        & 0.019                                                   & 0.034                                                      & 0.1               \\
Anonymization$\downarrow$              & FIT           & 0.042                                                   & 0.072                                                      &\underline{0.002}  \\
                                       & Deep Privacy  & 0.077                                                   & 0.092                                                      & 0.067             \\
                                       & Deep Privacy2 & 0.042                                                   & 0.066                                                      & \textbf{0}        \\
                                       & FALCO         & 0.04                                                    & 0.059                                                      & \textbf{0}        \\ \hline
\multicolumn{1}{c}{}                   & Ours          & \textbf{1}                                              & \textbf{1}                                                 & \textbf{1}        \\
\multicolumn{1}{c}{Recovery$\uparrow$} & RiDDLE        & \underline{0.996}                                       & \underline{0.998}                                          & \underline{0.998} \\
\multicolumn{1}{c}{}                   & FIT           & 0.964                                                   & 0.976                                                      & \textbf{1}        \\
\bottomrule
\end{tabular}
\label{table:1}
\end{table}

\begin{table*}[!t]
\centering
\caption{Quantitative evaluation (image quality) on CelebA-HQ. The best results are in \textbf{bold}, and the second best results are \underline{underlined}.}
\begin{tabular*}{\textwidth}{@{\extracolsep{\fill}} cccccccc|ccc @{}}
\toprule
\multirow{2}{*}{Method} & \multicolumn{7}{c|}{Anonymization} & \multicolumn{3}{c}{Recovery} \\ 
\cline{2-11} 
                    & FALCO  & RiDDLE             & CIAGAN & FIT                & Deep Privacy & Deep Privacy2     & Ours            & FIT                & RiDDLE & Ours            \\ 
\hline
MSE$\downarrow$     & 0.104  & 0.055              & 0.137  & \underline{0.015}  & 0.025        & 0.024             & \textbf{0.013}  & \underline{0.006}  & 0.046  & \textbf{0.004}  \\
PSNR$\uparrow$      & 15.949 & 18.801             & 15.233 & \underline{24.605} & 22.351       & 22.421            & \textbf{24.978} & \underline{28.275} & 19.632 & \textbf{29.535} \\
LPIPS$\downarrow$   & 0.092  & 0.07               & 0.107  & 0.062              & 0.04         & \underline{0.028} & \textbf{0.016}  & \underline{0.055}  & 0.188  & \textbf{0.027}  \\
FID$\downarrow$     & 29.93  & \underline{15.389} & 32.611 & 24.578             & 23.751       & 16.091            & \textbf{9.61}   & \underline{25.768} & 46.390 & \textbf{7.917}  \\ 
\bottomrule
\end{tabular*}
\label{table:2}
\end{table*}

\subsection{Recovery}

\begin{figure}[!t]
\centering
\includegraphics[width=1.0\columnwidth]{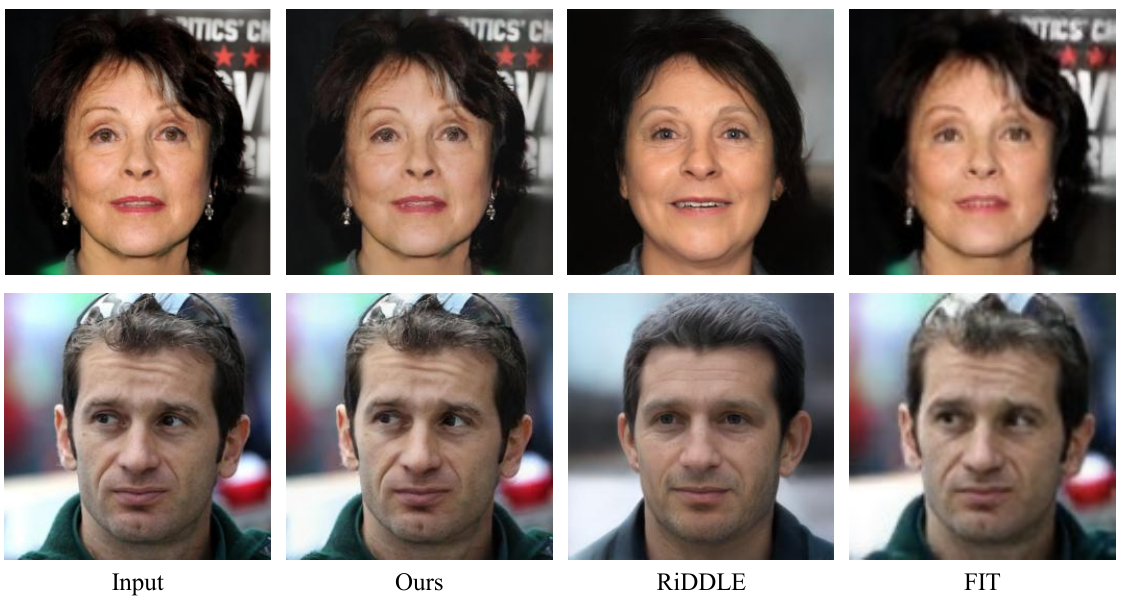}
\caption{Qualitative recovery results on FFHQ~\cite{karras2019style}. Our recovery results demonstrate exceptional accuracy and realism, while preserving image quality to a significant extent.}
\label{Fig6}
\end{figure}

\begin{figure}[!t]
\centering
\includegraphics[width=1.0\columnwidth]{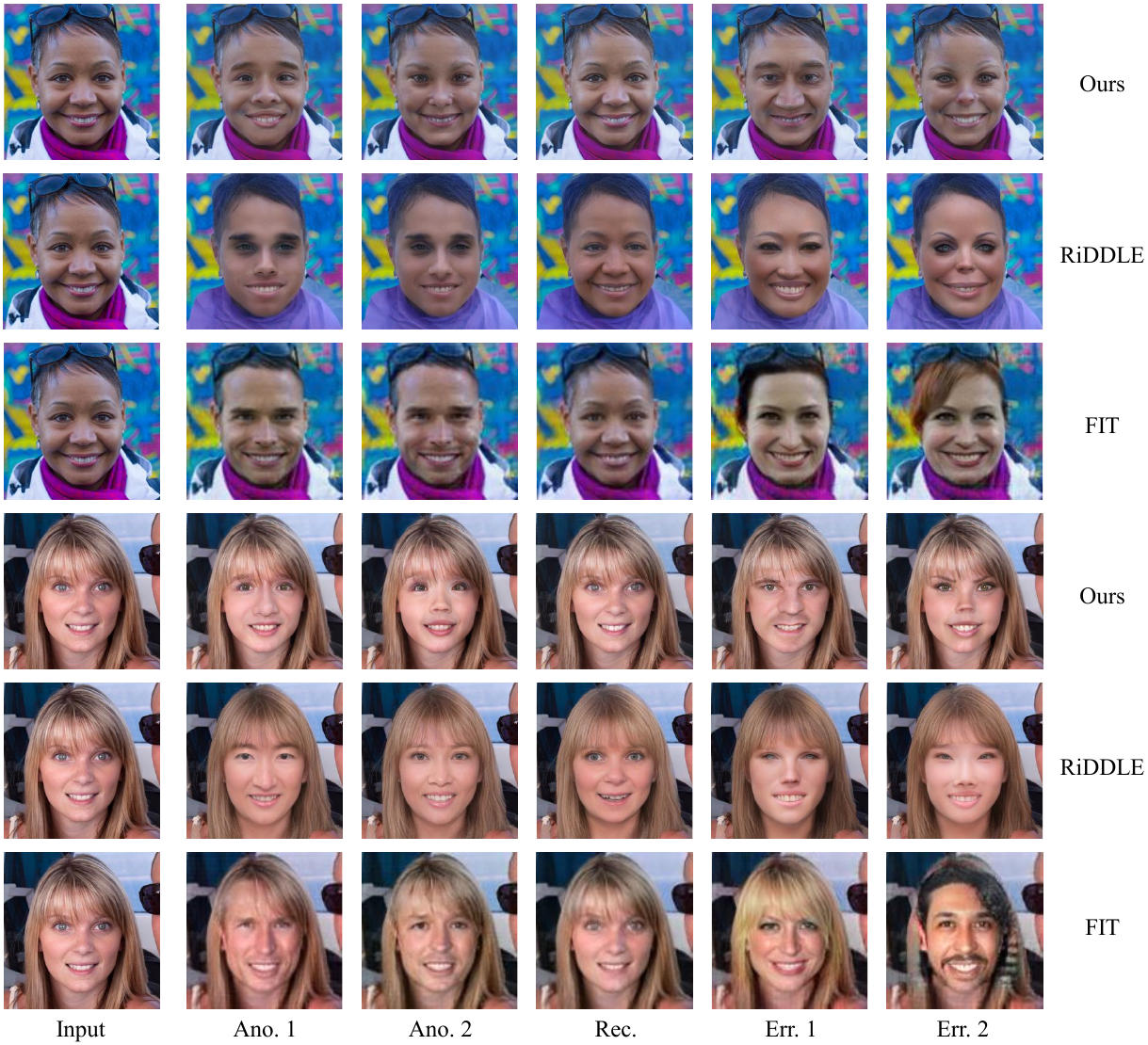}
\caption{Qualitative anonymization and recovery results on FFHQ~\cite{karras2019style}. Our method is capable of generating diverse, high-quality anonymization results while accurately recovering the original identity.}
\label{Fig7}
\end{figure}

\begin{figure}[!t]
\centering
\includegraphics[width=1.0\columnwidth]{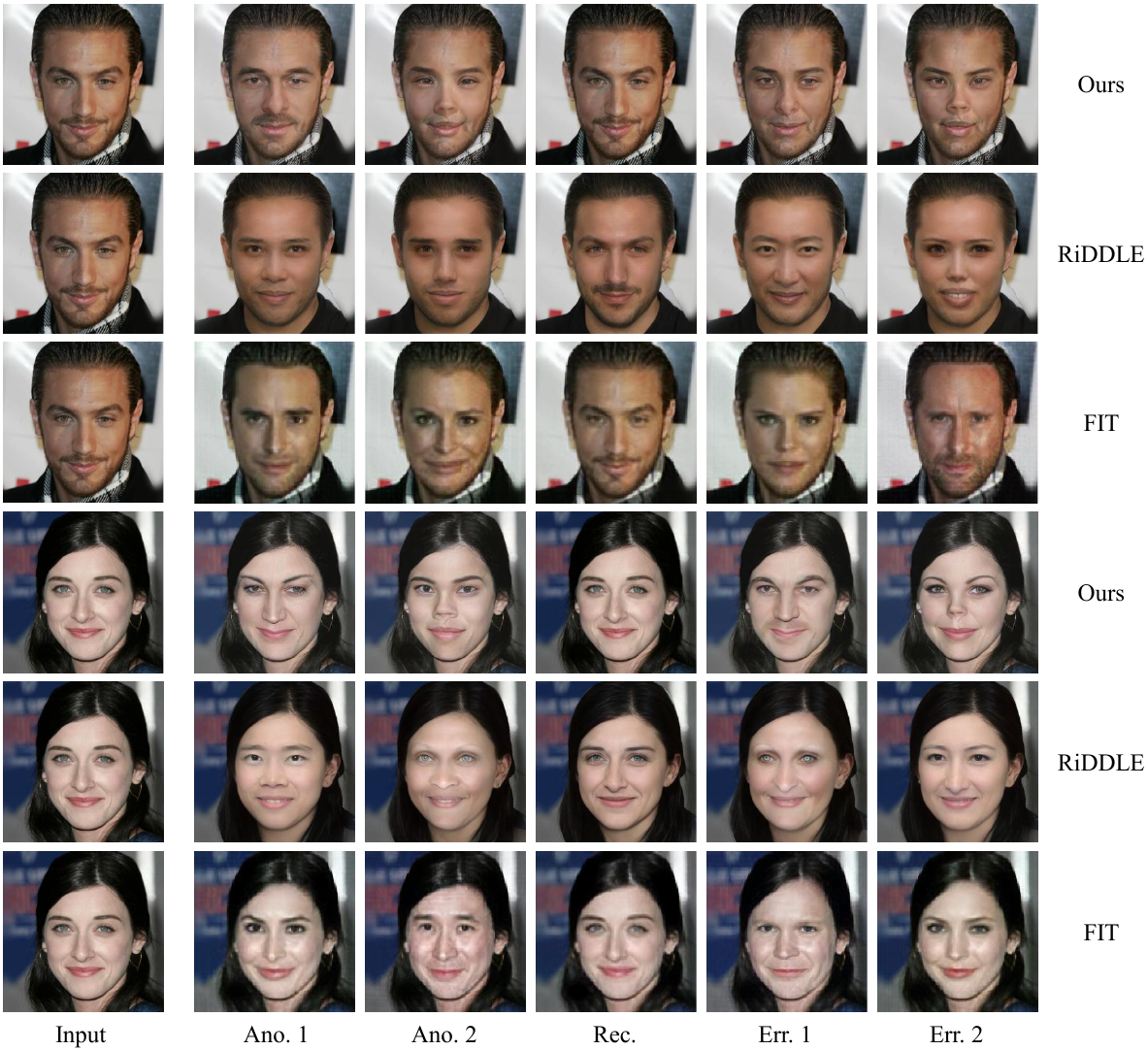}
\caption{Qualitative anonymization and recovery results on CelebA-HQ~\cite{karras2017progressive}. Our approach achieves superior visual quality compared to FIT~\cite{gu2020password} and RiDDLE~\cite{li2023riddle}}
\label{Fig8}
\end{figure}

We compare our recovery results on the FFHQ dataset with those generated by FIT~\cite{gu2020password} and RiDDLE~\cite{li2023riddle}, as illustrated in Fig.~\ref{Fig6}. Although the images recovered by RiDDLE largely preserve the original identity, both facial details and background information are significantly lost compared to the original images. Additionally, identity-independent facial features, such as expressions, are not retained. FIT, on the other hand, nearly restores the original identity accurately; however, the image quality deteriorates compared to the original, with noticeable blurring, particularly upon zooming in. In contrast, our recovery results exhibit exceptional accuracy and realism while maintaining superior image quality. According to our task objectives, different keys should produce distinct anonymization results during the anonymization process. During recovery, the correct key should restore the correct identity, while incorrect keys should fail to recover the correct identity. Moreover, the images generated using incorrect keys should also possess high quality to mislead attackers, making it difficult for them to determine whether the input key is incorrect. As clearly illustrated in Fig.~\ref{Fig7} and Fig.~\ref{Fig8}, our method effectively satisfies these requirements. In contrast, RiDDLE introduces unnatural shadows around the eyes during the anonymization and recovery process. For FIT’s incorrectly recovered results, attributes such as hair are altered, and visible stitching artifacts can be observed. Furthermore, FIT consistently produces images of lower quality.

The quantitative evaluation results are shown in TABLE~\ref{table:1} and TABLE~\ref{table:2}. A higher recognition rate for recovered images indicates better recovery performance. Our method surpasses existing reversible anonymization approaches in terms of identity recovery accuracy, while also delivering superior quality in the recovered images compared to current methods.

\subsection{Occlusion Robustness}
\begin{figure}[!t]
\centering
\includegraphics[width=1.0\columnwidth]{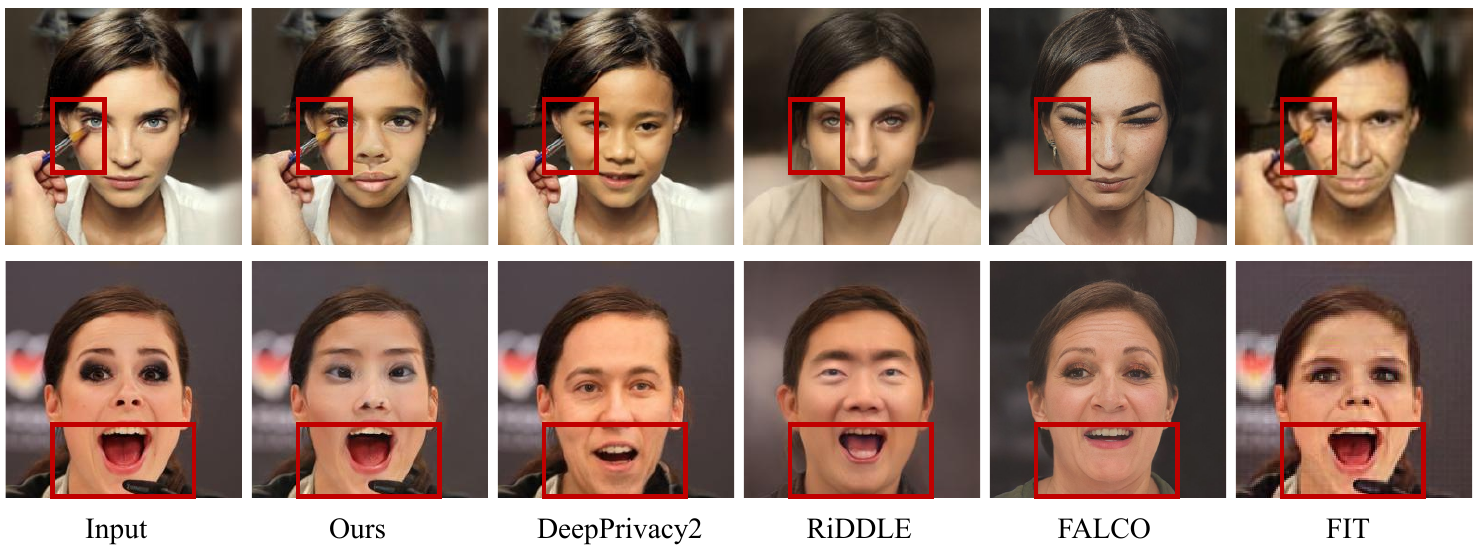}
\caption{De-identification results for faces with occlusions.}
\label{Fig9}
\end{figure}

Facial occlusion presents a significant challenge in the task of face de-identification. Existing methods struggle to achieve satisfactory anonymization and recovery results in scenarios where partial facial occlusion is present. As shown in Fig.~\ref{Fig9}, DeepPrivacy2 generates de-identified faces but fails to retain occlusions, and the resulting expressions deviate significantly from those in the original image. RiDDLE, while capable of anonymizing faces and preserving original expressions, struggles to maintain non-facial attributes like the background and occlusions. FALCO produces de-identified faces, but with expressions that differ markedly from the original, along with noticeable alterations to the hairstyle and the loss of occlusions. FIT retains occlusions and expressions but delivers low-quality images with significant blurring of details. In contrast, our approach consistently achieves effective de-identification while preserving occlusions, retaining original expressions, and producing results of superior quality and realism.

\begin{figure}[!t]
\centering
\includegraphics[width=1.0\columnwidth]{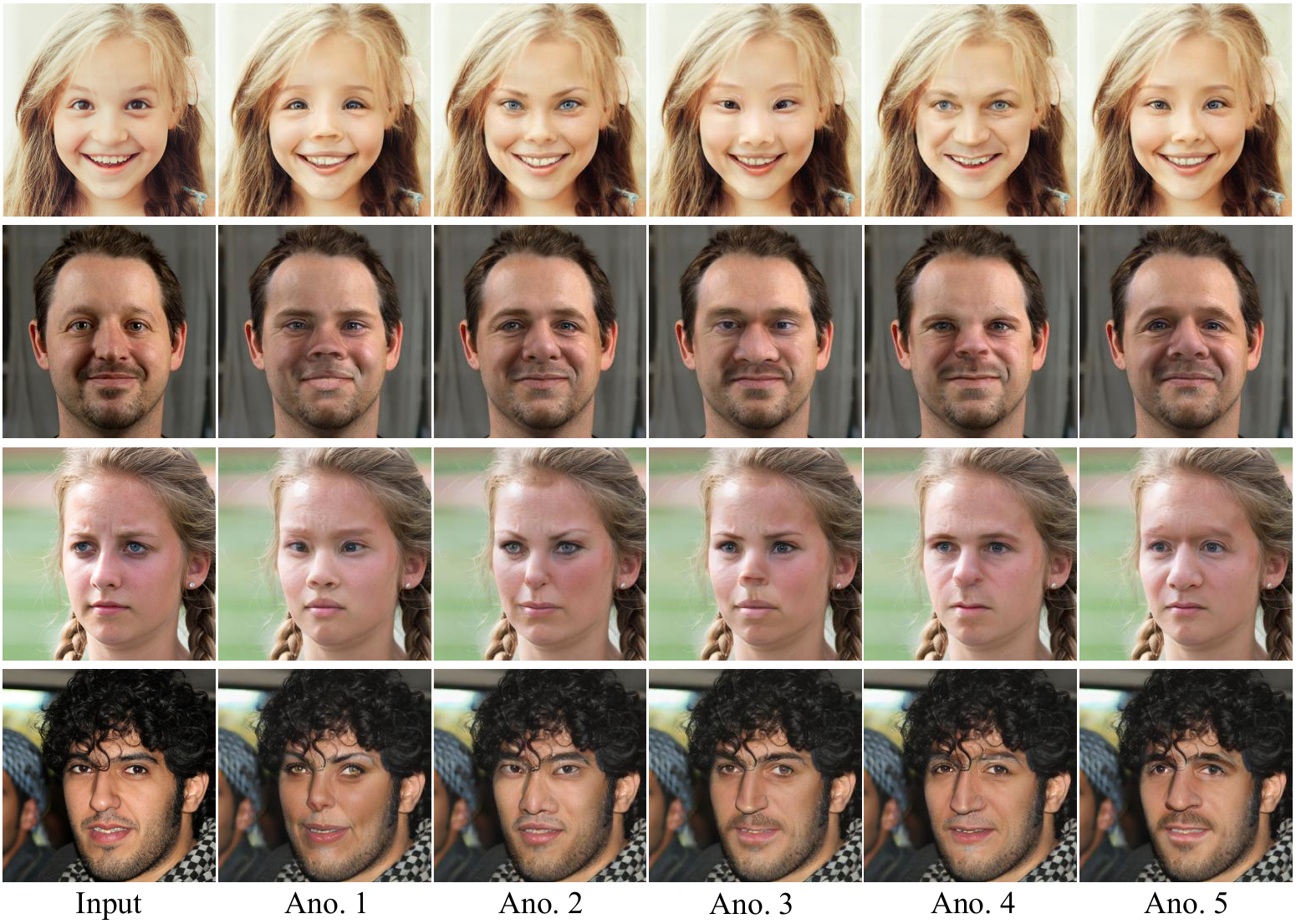}
\caption{Diverse anonymization results on FFHQ~\cite{karras2019style}. The first column shows the input images, while the remaining columns display anonymized images generated using various keys.}
\label{Fig10}
\end{figure}

\subsection{Diversity of Identities}
\begin{figure}[!t]
\centering
\includegraphics[width=1.0\columnwidth]{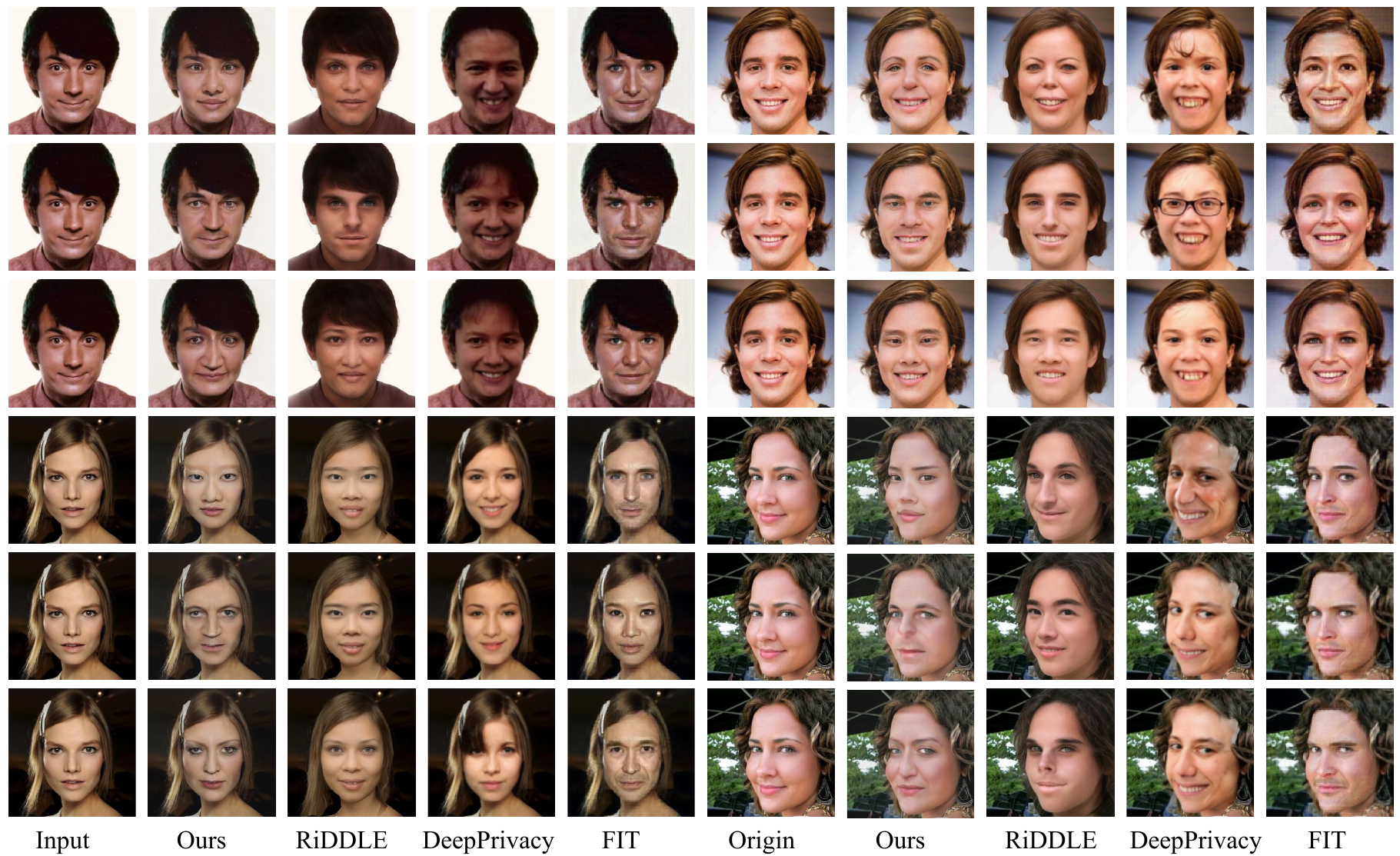}
\caption{Qualitative comparison of diversity with leading anonymization methods.}
\label{Fig11}
\end{figure}

Fig.~\ref{Fig10} illustrates the anonymization results on the FFHQ dataset using five distinct passwords. The de-identified images produced by our method exhibit significant identity variations, particularly in the eye and mouth regions. To evaluate the diversity of de-identified images, we compared our method against FIT, RiDDLE, and DeepPrivacy. The comparative results are presented in Fig.~\ref{Fig11}. FIT generates faces with similar features concentrated in specific local regions, limiting diversity. DeepPrivacy, on the other hand, often produces results with unnatural artifacts. RiDDLE fails to achieve diverse anonymization, as facial features remain largely unchanged across different passwords, resulting in highly similar anonymized outputs. In contrast, our method demonstrates superior diversity in de-identification results.

\begin{figure}[!t]
\centering
\subfigure[Ours (ArcFace)]{
    \begin{minipage}[h]{0.33\linewidth}
        \centering
        \includegraphics[width=1.2in]{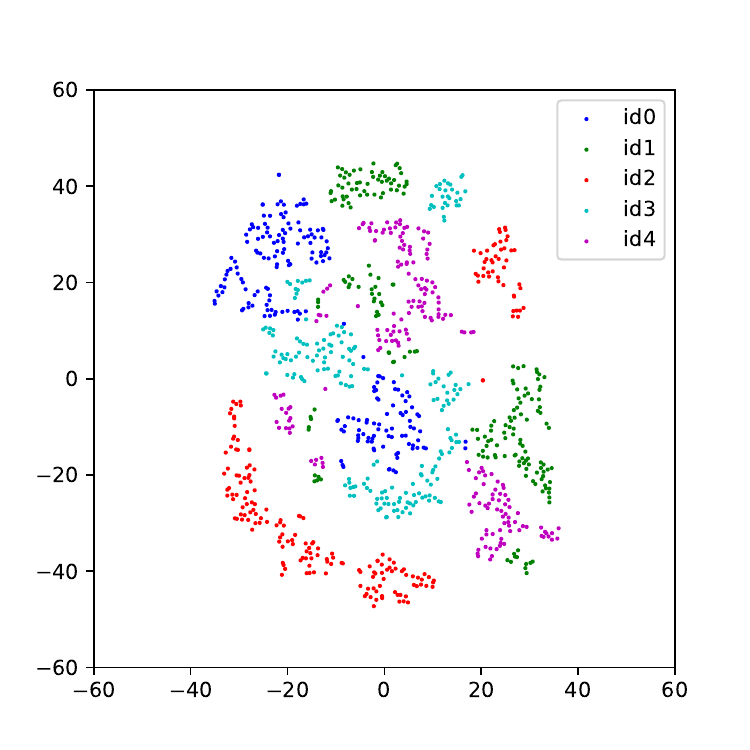}\\
        \vspace{0.02cm}
    \end{minipage}%
}%
\subfigure[RiDDLE (ArcFace)]{
    \begin{minipage}[h]{0.33\linewidth}
        \centering
        \includegraphics[width=1.2in]{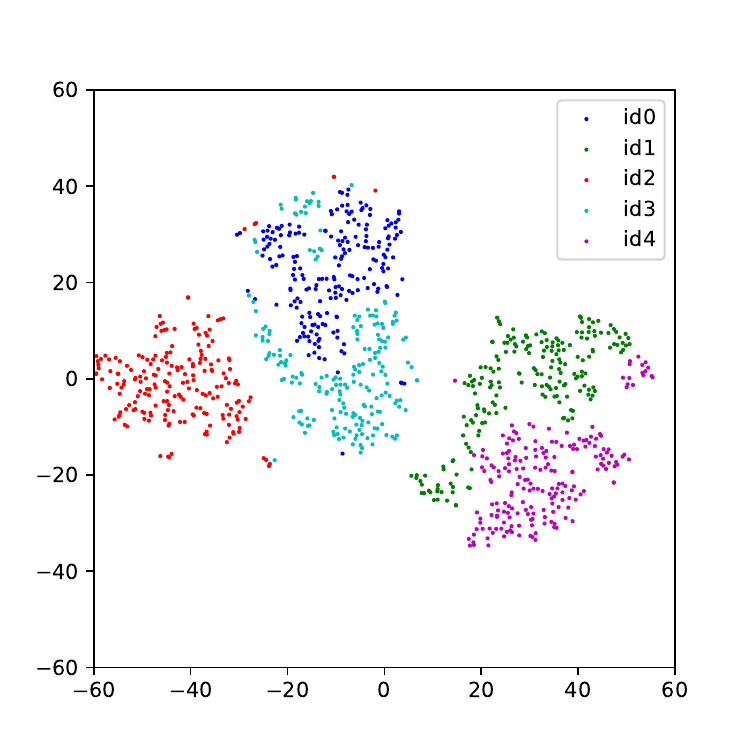}\\
        \vspace{0.02cm}
    \end{minipage}%
}%
\subfigure[FIT (ArcFace)]{
    \begin{minipage}[h]{0.33\linewidth}
        \centering
        \includegraphics[width=1.2in]{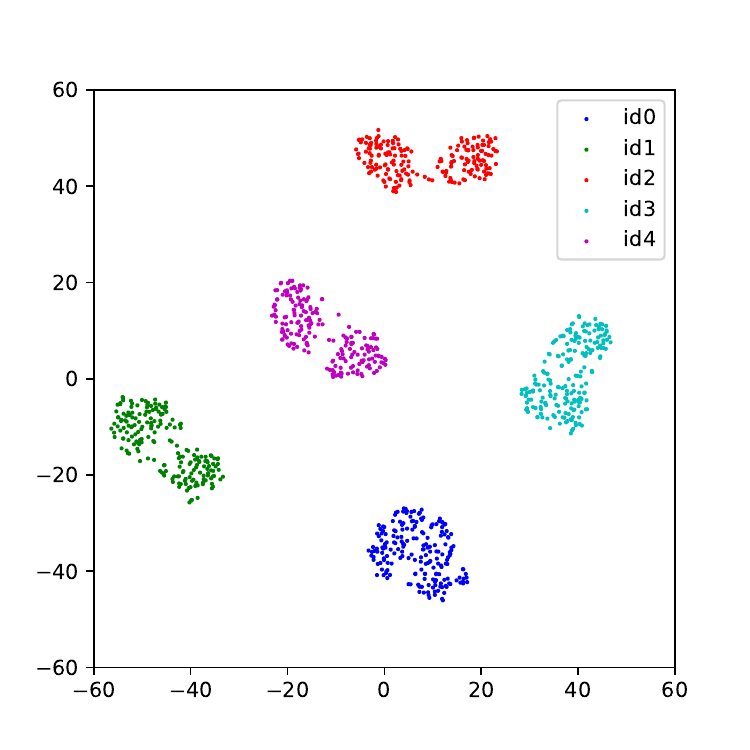}\\
        \vspace{0.02cm}
    \end{minipage}%
}%

\subfigure[Ours (FaceNet)]{
    \begin{minipage}[h]{0.33\linewidth}
        \centering
        \includegraphics[width=1.2in]{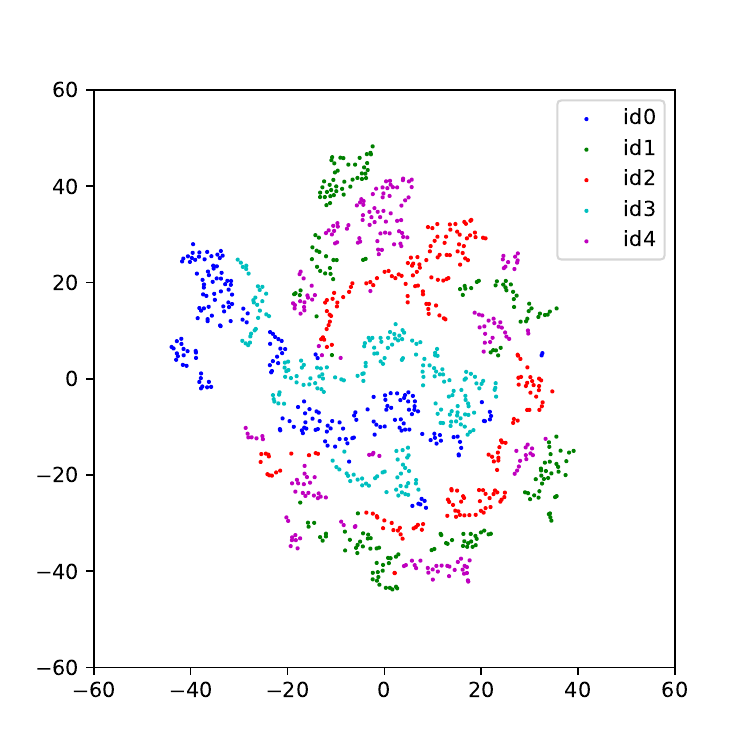}\\
        \vspace{0.02cm}
    \end{minipage}%
}%
\subfigure[RiDDLE (FaceNet)]{
    \begin{minipage}[h]{0.33\linewidth}
        \centering
        \includegraphics[width=1.2in]{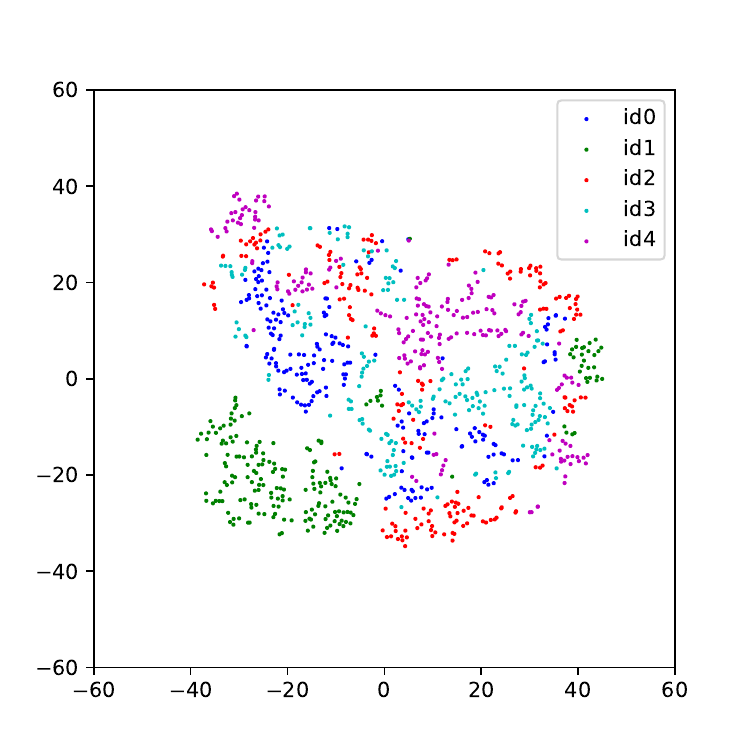}\\
        \vspace{0.02cm}
    \end{minipage}%
}%
\subfigure[FIT (FaceNet)]{
    \begin{minipage}[h]{0.33\linewidth}
        \centering
        \includegraphics[width=1.2in]{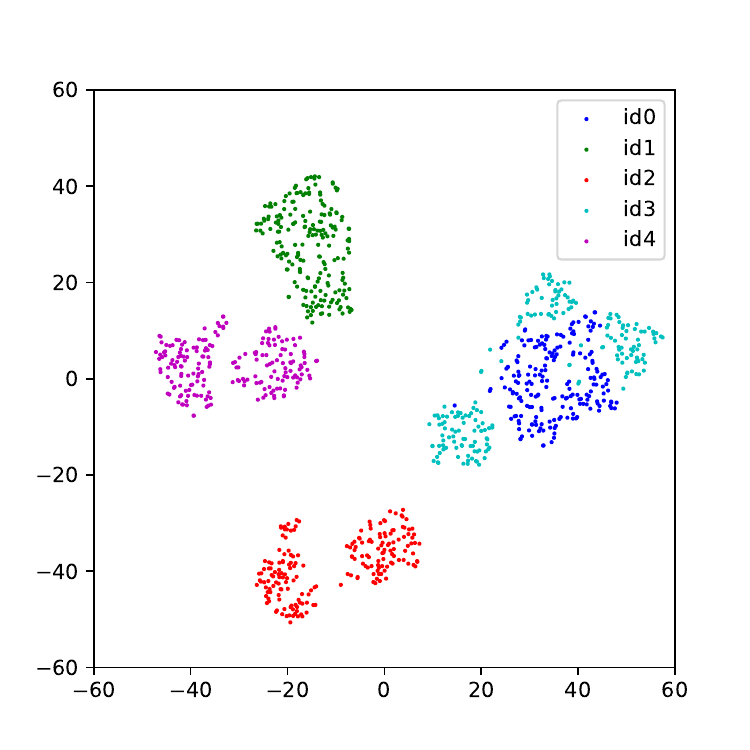}\\
        \vspace{0.02cm}
    \end{minipage}%
}%
\centering
\caption{The comparison of t-SNE~\cite{van2008visualizing} visualizations of identity embeddings in ArcFace and FaceNet with anonymization results from different methods. Points with the same color indicate that they share the same original identity. A greater dispersion of points with the same color in the feature space indicates higher diversity in the anonymized images.}
\vspace{-0.2cm}
\label{Fig12}
\end{figure}

To further illustrate the diversity of our approach, we conduct identity visualization experiments. Five facial images are randomly selected from the CelebA-HQ dataset, and 200 unique passwords are used to generate their de-identified versions. Identity embeddings of the anonymized images are extracted using the ArcFace~\cite{deng2019arcface} and FaceNet~\cite{schroff2015facenet} face recognition networks. These embeddings are then visualized in reduced dimensions using the t-SNE algorithm~\cite{van2008visualizing}. As shown in Fig.~\ref{Fig12}, the visualization results using ArcFace reveal that identity clusters for FIT and RiDDLE are tightly grouped, with large separations between clusters in the hyperplane. In contrast, our method produces more scattered identity distributions, occupying a broader region of the hyperplane. Similarly, with FaceNet, FIT's identity clusters remain tightly grouped, and some clusters in RiDDLE (\textit{e.g.}, the green identity) are also relatively compact. However, our de-identified faces are again more widely distributed, occupying most of the hyperplane. These results highlight the greater diversity achieved by our method compared to existing approaches.

\begin{table*}[!t]
\centering
\caption{Utility evaluation of de-identified results. The best results are in \textbf{bold}, and the second best results are \underline{underlined}.}
\begin{tabular*}{\textwidth}{@{\extracolsep{\fill}} cc|ccccccc @{}}
\toprule
\multicolumn{2}{c|}{Method}                                              & Ours   & RiDDLE            & CIAGAN      & FIT                & DeepPrivacy         & DeepPrivacy2       & FALCO                       \\ \hline
\multicolumn{1}{c|}{\multirow{2}{*}{Face detection$\uparrow$}}           & MtCNN  & \textbf{1}        & \textbf{1}  & \underline{0.997}  & \textbf{1}          & \textbf{1}         & \textbf{1}    & \textbf{1}  \\
\multicolumn{1}{c|}{}                                                    & Dlib   & \textbf{1}        & \textbf{1}  & 0.953              & 0.993               & \underline{0.995}  & \textbf{1}    & \textbf{1}  \\ \hline
\multicolumn{1}{c|}{\multirow{2}{*}{Bounding box distance$\downarrow$}}  & MtCNN  & \textbf{4.972}    & 6.164       & 20.863             & \underline{5.663}   & 6.131              & 6.212         & 7.882       \\
\multicolumn{1}{c|}{}                                                    & Dlib   & \textbf{3.905}    & 5.201       & 19.324             & 4.489               & \underline{4.351}  & 5.463         & 6.314       \\ \hline
\multicolumn{1}{c|}{\multirow{2}{*}{Landmark distance$\downarrow$}}      & MtCNN  & \underline{2.540} & 3.619       & 7.568              & \textbf{2.379}      & 4.353              & 6.021         & 3.913       \\
\multicolumn{1}{c|}{}                                                    & Dlib   & \textbf{2.238}    & 3.658       & 9.679              & \underline{2.639}   & 4.116              & 5.842         & 4.123       \\
\bottomrule
\end{tabular*}
\label{table:3}
\end{table*}

\begin{table*}[!t]
\centering
\caption{Quantitative ablation results. The best results are in \textbf{bold}, and the second best results are \underline{underlined}.}
\begin{tabular*}{\textwidth}{@{\extracolsep{\fill}} ccccc|cccc @{}}
\toprule
\multirow{2}{*}{Configuration} & \multicolumn{4}{c|}{Anonymization} & \multicolumn{4}{c}{Recovery} \\ 
\cline{2-9} 
                        & Single-stage       & w/o CID            & w/o MAAR  & Full             & Single-stage  & w/o CID             & w/o MAAR  & Full             \\   \hline
MSE$\downarrow$         & 0.033              & \underline{0.015}  & 0.129     & \textbf{0.013}   & 0.014         & \underline{0.005}   & 0.117     & \textbf{0.004}   \\
LPIPS$\downarrow$       & 0.024              & 0.02               & 0.096     & \textbf{0.016}   & 0.032         & 0.03                & 0.09      & \textbf{0.027}   \\
PSNR$\uparrow$          & 21.658             & \underline{24.515} & 15.139    & \textbf{24.978}  & 25.785        & \underline{28.542}  & 15.574    & \textbf{29.535}  \\
FID$\downarrow$         & 16.375             & \underline{11.07}  & 39.411    & \textbf{9.61}    & 9.919         & \underline{8.238}   & 35.534    & \textbf{7.917}   \\
De-identity$\downarrow$ & \underline{0.031}  & 0.035              & 0.035     & \textbf{0.02}    & —             & —                   & —        & —                 \\
Recovery$\uparrow$      & —                  & —                  & —         & —                & \textbf{1}    & \underline{0.99}    & 0.965     & \textbf{1}       \\ 
\bottomrule
\end{tabular*}
\label{table:4}
\end{table*}

\subsection{Image Utility}
The outcome of face de-identification should prevent attackers from detecting image edits while still allowing the image to be processed by various intelligent models. To assess this, we quantitatively evaluate the utility of de-identified images in downstream visual tasks. Specifically, we calculate the face detection rate, per-pixel distance of facial bounding boxes, and the distance between 68 facial landmarks using two face detection models: MtCNN \cite{zhang2016joint} and Dlib \cite{kazemi2014one}. This allows us to compare our method with other anonymization approaches. As shown in TABLE~\ref{table:3}, when evaluated using the Dlib model, our method achieves exceptional results in face detection rate, per-pixel distance, and the distance of 68 facial landmarks. Similarly, with the MtCNN model, our method performs exceptionally well in face detection rate and per-pixel distance, and achieves results comparable to FIT in terms of the 68 facial keypoint distance. These results demonstrate that our method preserves the utility of de-identified images, making it more effective for downstream visual tasks compared to other anonymization methods.

\begin{figure}[!t]
\centering
\includegraphics[width=1.0\columnwidth]{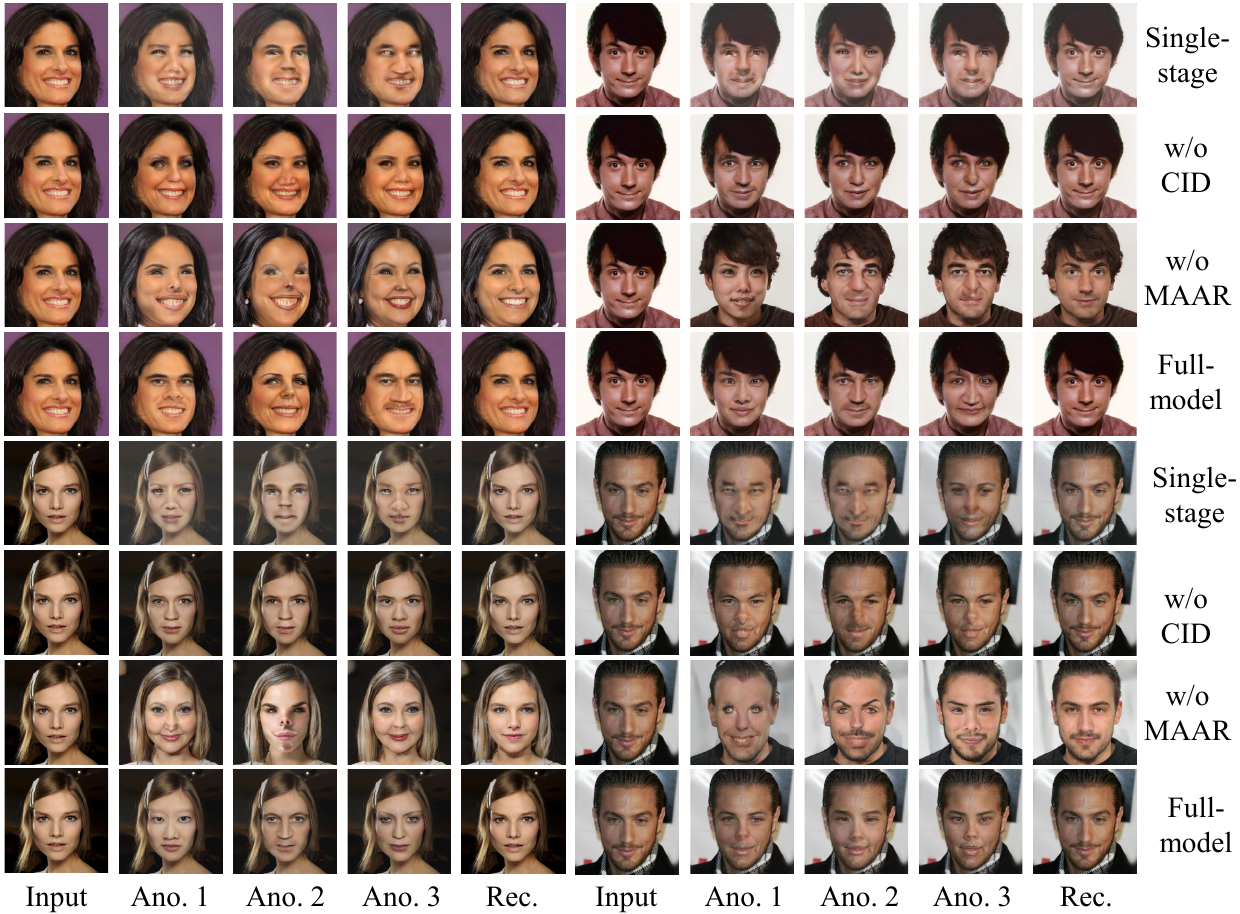}
\caption{Qualitative ablation results.}
\label{Fig13}
\end{figure}

\subsection{Ablation Study}
In this section, we qualitatively and quantitatively evaluate the contributions of key components in DBAF. Our primary contribution is the ``Disentangle before Anonymization'' strategy. Therefore, we first keep the overall network structure unchanged and directly perform disentanglement and anonymization in a single stage to test the performance under this setup. We refer to this ablation as ``single-stage''. The CID module contributes to disentangle identity and attribute. To evaluate its effectiveness, we set up an ablation configuration named ``w/o CID'', where the CID is replaced with a simple structure consisting of a four-layer MLP. The MAAR module contributes to preserve attribute-related details and enhance the network’s robustness against occlusion. To evaluate its effectiveness, we set up an ablation configuration named ``w/o MAAR'', where the MAAR module is removed during training. ``Full-model'' represents the complete DBAF.

The quantitative ablation results are presented in TABLE~\ref{table:4}. Across all metrics, the "Full-model" achieves the best performance, indicating that all modules in DBAF contribute to its overall effectiveness. The qualitative ablation results are shown in Fig.~\ref{Fig13}. The de-identified faces generated by the ``Single-stage'' configuration exhibit blurriness, artifacts, and other distortions. The ``w/o CID'' configuration produces de-identified faces with highly similar facial features, highlighting the effectiveness of CID in disentangling identity and attributes. The ``w/o MAAR'' configuration fails to preserve identity-independent attributes, such as background and hairstyle. In contrast, the ``Full-model'' generates more realistic and higher-quality de-identified images. 

\section{Conclusion}

This paper addresses the critical challenges in face de-identification by proposing a novel ``Disentangle before Anoymize'' framework. Our method overcomes the limitations of simultaneous training, enabling better preservation of attribute details and achieving high-quality anonymization. The integration of the CID and KRIA modules ensures faithful identity anonymization while maintaining attribute integrity, and the introduction of the MAAR module enhances robustness against occlusions, reducing editing artifacts. Extensive experimental results confirm the superiority of our approach over state-of-the-art methods, demonstrating its ability to produce authentic, attribute-preserved, and occlusion-robust anonymized results. In future work, we will explore a more compact identity disentanglement space to achieve more effective anonymization.

\bibliographystyle{IEEEtran}
\normalem
\bibliography{./Ref.bib}

\begin{IEEEbiography}[{\includegraphics[width=1in,height=1.25in,clip,keepaspectratio]{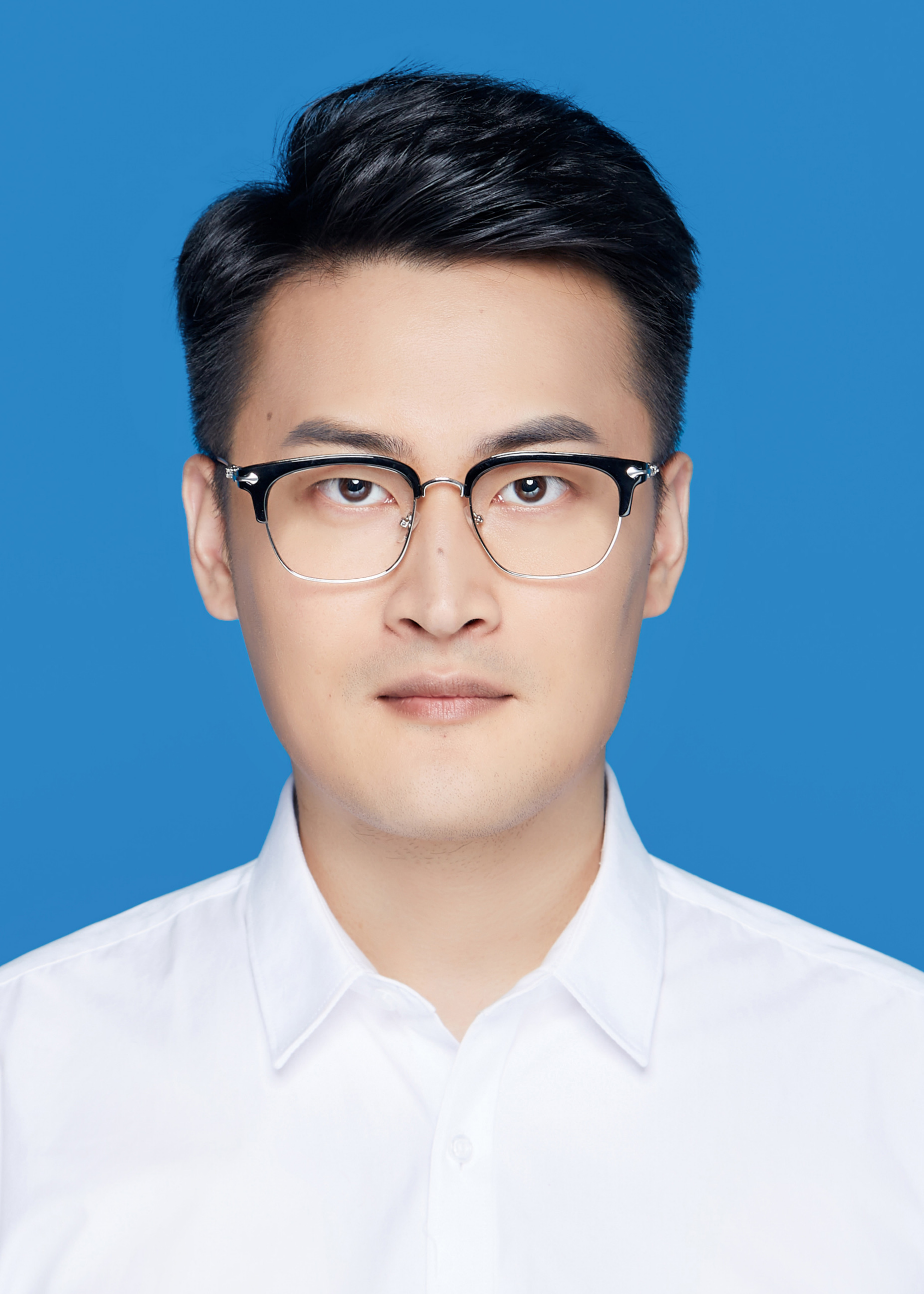}}]{Mingrui Zhu} received the B.Eng. degree in electronic and information engineering from Guangxi University in 2014 and the Ph.D. degree in circuits and systems from Xidian University in 2020. He is currently an Associate Professor with the State Key Laboratory of Integrated Services Networks, Xidian University, Xi'an, China. His current research interests include computer vision and machine learning.
\end{IEEEbiography}

\begin{IEEEbiography}[{\includegraphics[width=1in,height=1.25in,clip,keepaspectratio]{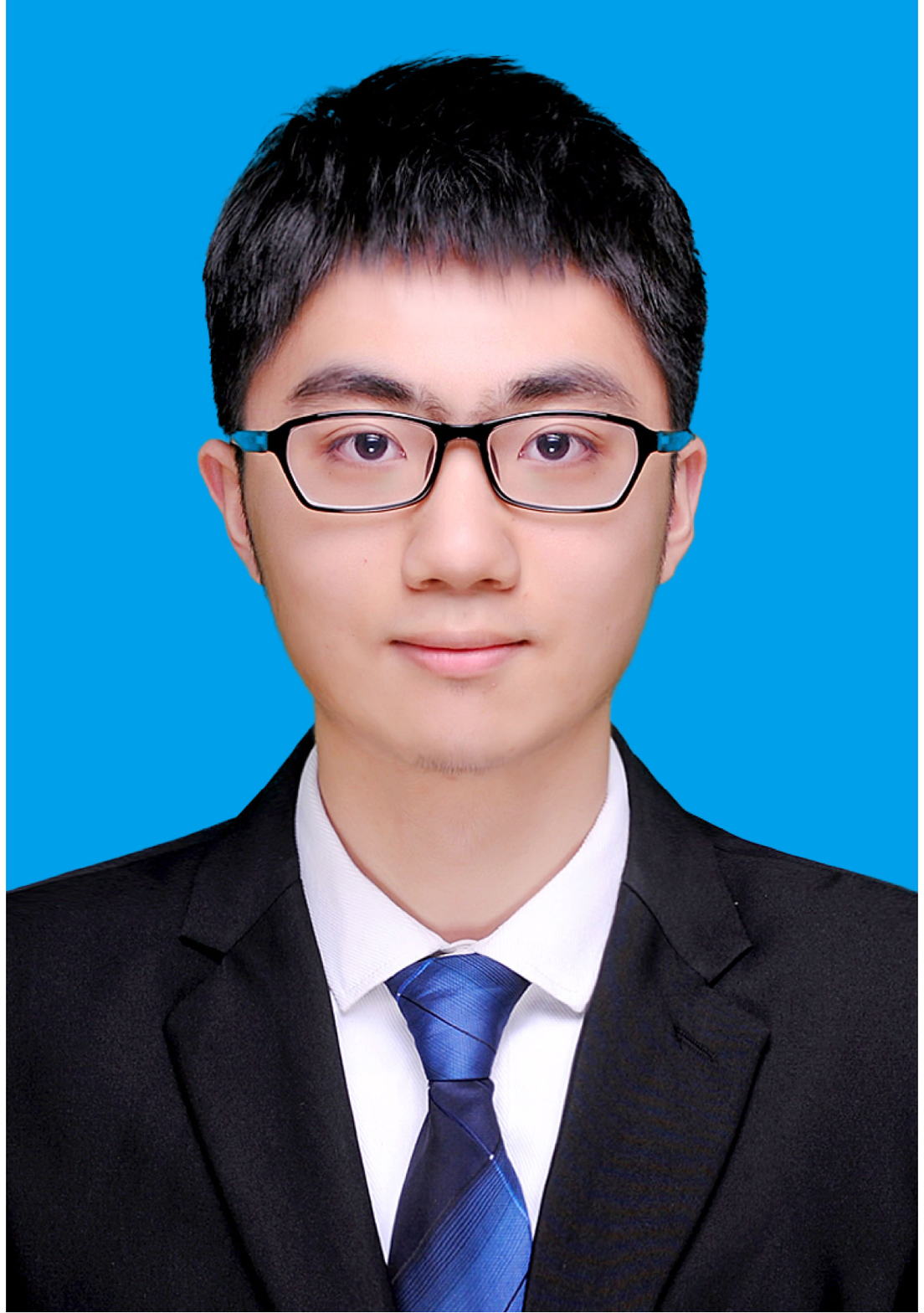}}]{Dongxin Chen} received the B.Eng. degree in Measurement and Control Technology and Instrument from Xidian University in 2022. He is currently pursuing the M.Sc. degree at Xidian University, Xi'an, China. His current research interests include computer vision and machine learning.
\end{IEEEbiography}

\begin{IEEEbiography}[{\includegraphics[width=1in,height=1.25in,clip,keepaspectratio]{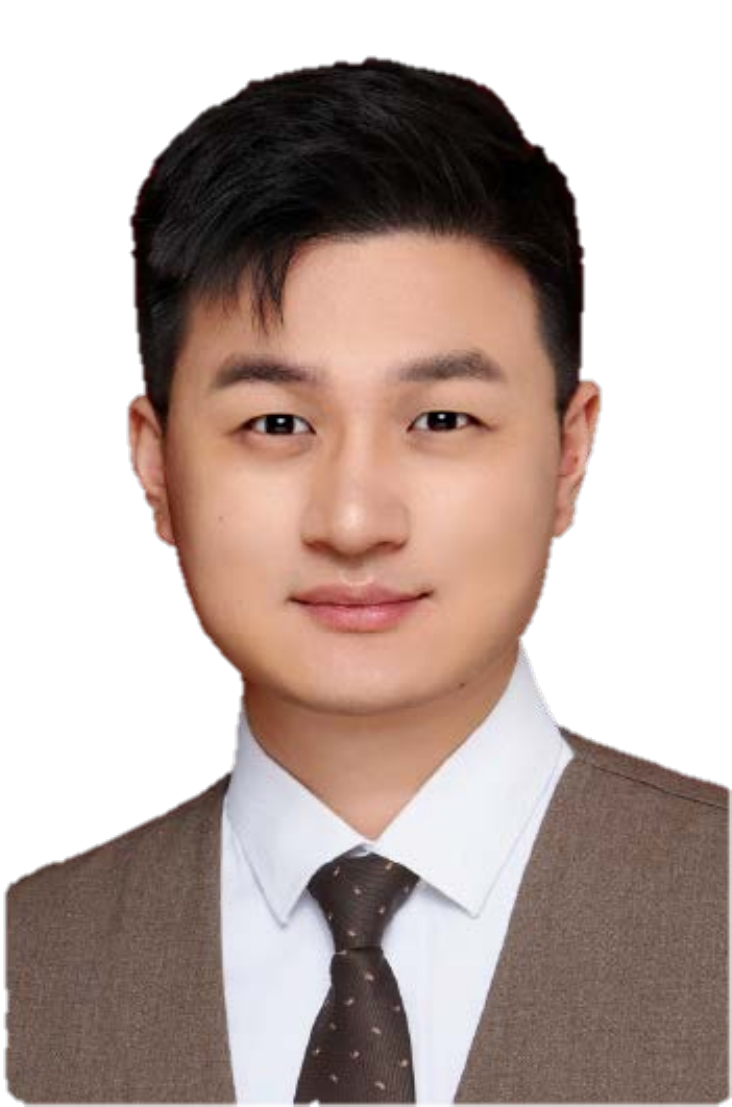}}]{Xin Wei} received the B.Sc. degree in pure mathematics from Wuhan University in 2016 and the Ph.D. degree in applied mathematics from Xi'an Jiaotong University, Xi'an, China, in 2024. He is currently an Associate Professor with the State Key Laboratory of Integrated Services Networks, Xidian University, Xi'an, China. His research interests include 3D shape analysis, LiDAR-based 3D recognition, and 3D domain generalization and adaptation.
\end{IEEEbiography}

\begin{IEEEbiography}[{\includegraphics[width=1in,height=1.25in,clip,keepaspectratio]{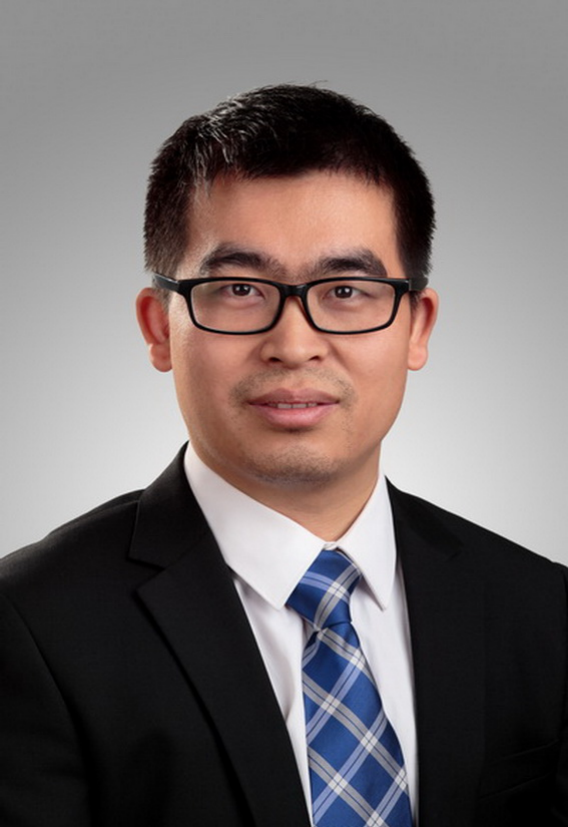}}]{Nannan Wang} (Senior Member, IEEE) received the B.Sc. degree in information and computation science from the Xi'an University of Posts and Telecommunications in 2009 and the Ph.D. degree in information and telecommunications engineering from Xidian University in 2015. He is currently a Professor with the State Key Laboratory of Integrated Services Networks, Xidian University, Xi'an, China. He has published over 150 articles in refereed journals and proceedings, including IEEE T-PAMI, IJCV, CVPR, ICCV, etc. His current research interests include computer vision and machine learning. 
\end{IEEEbiography}

\begin{IEEEbiography}[{\includegraphics[width=1in,height=1.25in,clip,keepaspectratio]{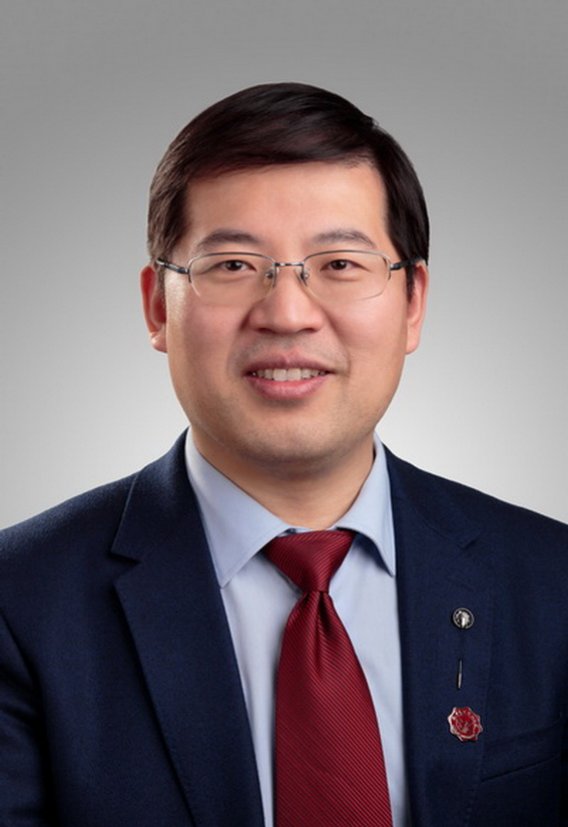}}]{Xinbo Gao} (Fellow, IEEE) received the B.Eng., M.Sc., and Ph.D. degrees in electronic engineering, signal and information processing from Xidian University, Xi’an, China, in 1994, 1997, and 1999, respectively. From 1997 to 1998, he was a Research Fellow with the Department of Computer Science, Shizuoka University, Shizuoka, Japan. From 2000 to 2001, he was a Post-Doctoral Research Fellow with the Department of Information Engineering, The Chinese University of Hong Kong, Hong Kong. Since 2001, he has been with the School of Electronic Engineering, Xidian University. He is also a Cheung Kong Professor of the Ministry of Education of China, a Professor of pattern recognition and intelligent system with Xidian University, and a Professor of computer science and technology with the Chongqing University of Posts and Telecommunications. He has published six books and around 300 technical articles in refereed journals and proceedings. His current research interests include image processing, computer vision, multimedia analysis, machine learning, and pattern recognition. He is a fellow of the Institute of Engineering and Technology and the Chinese Institute of Electronics. He served as the General Chair/Co-Chair, the Program Committee Chair/Co-Chair, or a PC Member for around 30 major international conferences. He is on the Editorial Boards of several journals, including \emph{Signal Processing} (Elsevier) and \emph{Neurocomputing} (Elsevier).
\end{IEEEbiography}

\end{document}